\documentclass[preprint,12pt,3p]{elsarticle}

\usepackage{graphicx}
\usepackage{amsmath}
\usepackage{amsfonts}
\usepackage{multirow}
\usepackage{caption}
\usepackage{subcaption}
\usepackage{booktabs}
\usepackage{makecell}
\usepackage[table]{xcolor}
\usepackage{upgreek}
\DeclareMathAlphabet{\mathcal}{OMS}{cmsy}{m}{n}
\SetMathAlphabet{\mathcal}{bold}{OMS}{cmsy}{b}{n}

\usepackage{color}
\definecolor{DarkGreen}{rgb}{0.1,0.9,0.1} % to color links in references
\usepackage[%
    pdftex,%
    colorlinks=true,%
    hyperindex,%
    plainpages=false,%
    pagebackref=true,%
    bookmarksopen,%
    bookmarksnumbered %
      ]{hyperref}

\newcommand{\hl}[1]{\cellcolor[HTML]{D6DBDF}\textbf{#1}}

\makeatletter
\AtBeginDocument{\def\@citecolor{DarkGreen}}
\makeatother

\makeatletter
\def\ps@pprintTitle{%
   \let\@oddhead\@empty
   \let\@evenhead\@empty
   \let\@oddfoot\@empty
   \let\@evenfoot\@oddfoot
}
\makeatother

\journal{Transportation Research Part C: Emerging Technologies}

\begin{document}

\begin{frontmatter}

\title{Graph Markov Network for Traffic Forecasting with Missing Data}

%%\author[label1]{Zhiyong Cui\fnref{label3}}
\author[label1]{Zhiyong Cui}
\address[label1]{Department of Civil and Environmental Engineering, University of Washington, Seattle, WA 98195 USA}
%%\fntext[label3]{I also want to inform about\ldots}
\ead{zhiyongc@uw.edu}
%% \ead[url]{zhiyongc.github.io}

\author[label2]{Longfei Lin}
\address[label2]{School of Electronic and Information Engineering, Beihang University, 100191 Beijing, China}
\ead{linlongfei9858@buaa.edu.cn}

\author[label1]{Ziyuan Pu}
\ead{ziyuanpu@uw.edu}

\author[label1]{Yinhai Wang\corref{cor1}}
\cortext[cor1]{Corresponding author}
\ead{yinhai@uw.edu}

\begin{abstract}
Traffic forecasting is a classical task for traffic management and it plays an important role in intelligent transportation systems. However, since traffic data are mostly collected by traffic sensors or probe vehicles, sensor failures and the lack of probe vehicles will inevitably result in missing values in the collected raw data for some specific links in the traffic network. Although missing values can be imputed, existing data imputation methods normally need long-term historical traffic state data. As for short-term traffic forecasting, especially under edge computing and online prediction scenarios, traffic forecasting models with the capability of handling missing values are needed. In this study, we consider the traffic network as a graph and define the transition between network-wide traffic states at consecutive time steps as a graph Markov process. In this way, missing traffic states can be inferred step by step and the spatial-temporal relationships among the roadway links can be Incorporated. Based on the graph Markov process, we propose a new neural network architecture for spatial-temporal data forecasting, i.e. the graph Markov network (GMN). By incorporating the spectral graph convolution operation, we also propose a spectral graph Markov network (SGMN). The proposed models are compared with baseline models and tested on three real-world traffic state datasets with various missing rates. Experimental results show that the proposed GMN and SGMN can achieve superior prediction performance in terms of both accuracy and efficiency. Besides, the proposed models' parameters, weights, and predicted results are comprehensively analyzed and visualized.
\end{abstract}


\end{frontmatter}

% \linenumbers


\section{Introduction}
Traffic forecasting, as a challenging topic for both academia and industry, has been under active research, development, and implementation for more than 40 years \cite{lana2018imputation}. Traffic forecasting plays an important role in transportation management and the general planning process. With the exponential increase in the volume of traffic data and the computational capability, traffic forecasting methods have been gradually shifting from classical statistical models to data-driven machine learning-based methods \cite{vlahogianni2014short}. In recent years, the rise of artificial intelligence (AI), especially deep learning methods, has dramatically stimulated the traffic forecasting research field. By leveraging the spatial-temporal patterns contained in immense data resources, many deep neural network models, including recurrent neural network (RNN), convolutional neural network (CNN), generative adversarial network (GAN), etc., have been widely applied in traffic forecasting studies and achieved state-of-the-art prediction performance. 

However, since network-wide traffic state data are mostly collected by traffic sensors or probe vehicles, sensor failures or irregular sampling from probe vehicles will result in missing values in the collected data. The missing value issue usually leads to an apparent decline in the forecasting performance, as most of the existing methods for traffic forecasting are not capable of dealing with missing values. Thus, forecasting performance of the models that only accept valid data as input will be significantly affected and limited. 

The regular solution for the missing data issue is to conduct data imputation, which targets to estimate the corrupted or missing traffic data. Due to the complex spatial-temporal patterns of traffic data, existing novel and effective data imputation methods, such as the Bayesian tensor decomposition approach \cite{chen2019bayesian} and the generative adversarial imputation network \cite{yoon2018gain}, usually need large datasets covering a long period of time to achieve good imputation performance. However, a large dataset covering a long period of time is not always available. Further, in the connected vehicle environments or under the edge computing scenarios, online forecasting computations need to be completed in devices with limited storage and computational capabilities. Thus, those solutions requiring large datasets is not feasible for real-time traffic forecasting tasks. 

To solve the missing value issue and fulfill traffic forecasting at the same time, traffic forecasting models with the capability of dealing with missing values have also been proposed. However, most of the existing models \cite{zhang2016comparative,lee1999application}, take the spatial-temporal traffic data as multivariate time series, and thus, they neglect the important spatial influence between the road links in the traffic network. There are several deep learning-based methods \cite{duan2016efficient,tian2018lstm} taking spatial factors into consideration. However, they still cannot incorporate the intrinsic structure of the traffic network into the traffic forecasting process. 

In this study, to overcome the problems mentioned above, we propose a graph Markov network (GMN), which is a new neural network architecture for spatial-temporal data forecasting with missing values. The traffic network is converted into a graph with topological properties. We consider the variations of the traffic states in the traffic network has Markov property and graph localization property. Based on the two properties, the traffic state transition process can be considered as a graph Markov process. The GMN is designed based on the graph Markov process, which inherently incorporates the spatial-temporal relationships among the links in the traffic network. By incorporating the spectral graph convolution operation, we also propose a spectral graph Markov network (SGMN). Experimental results indicate that the proposed SGMN and GMN can achieve superior prediction performance with greater efficiency.

The contributions of this study can be summarized as follows:
\begin{enumerate}
    \item We consider the traffic network as a graph and define the transition between network-wide traffic states at consecutive time steps as a graph Markov process. 
    \item We propose a new neural network structure, i.e. the graph Markov network, based on the proposed graph Markov process for dealing with missing values and forecasting traffic state simultaneously.
    \item By incorporating the spectral graph convolution operation, we also propose a spectral graph Markov network. 
    \item Experiment results tested on three real-world network-wide traffic state datasets show that the proposed models can achieve superior prediction performance in terms of both accuracy and efficiency.
\end{enumerate}

The rest of this paper is organized as follows: the second section describes the related studies on traffic forecasting with missing values. The third section introduces the proposed graph Markov process and the proposed GMN model. The fourth section discusses the experimental results and the concluding remarks are presented in the fifth section.
% However, since the traffic sensors may fail and the 

% has been studied and classified into multiple categories
% introduce traffic prediction \\

% existing problems: missing value \\

% solutions: data imputation  \\

% another problem: real-time prediction with missing values \\

% existing methods: GRU-D LSTM-M \\

% problems: they are RNN structures, which consider the historical data as a time sequence. \\

% we propose a method from another perspective: consider the dynamic of traffic network as a Markov process. current state is determined only by the previous time step.  \\

% incorporating graph  \\

% It is totally a new structure to handle missing values in the traffic prediction problem. \\

% our contribution \\

% the rest of the paper organization \\

\section{Literature Review}
% \subsection{Classical Traffic Forecasting Models}
% \subsection{Deep Learning Models for Traffic Forecasting}
% \subsection{Deep Learning Models for Traffic Forecasting with missing values}

Classical traffic forecasting models can generally be classified into two categories, traditional statistical models and computational intelligence, i.e. machine learning-based, models \cite{vlahogianni2014short}. The statistical methods are mostly parametric approaches, including variants of auto-regressive integrated moving average (ARIMA) models \cite{williams2001multivariate}, parametric Kalman filtering models \cite{okutani1984dynamic}, and other types of time-series models \cite{ghosh2009multivariate}, that are developed based on a predefined model structure with theoretical assumptions and the parameters are calibrated using historical data \cite{smith2002comparison}. With the ability to accommodate the stochastic and non-linear nature of traffic patterns, classical machine learning methods are widely adopted for the traffic forecasting task, such as support vector regression \cite{wu2004travel}, Bayesian network approaches \cite{sun2006bayesian}. In recent years, with the rise of AI, the performance of emerging deep learning-based traffic forecasting methods outperform that of classical methods.

\subsection{Deep learning-based traffic forecasting methods} 
Since the traffic data contain both spatial and temporal attributes, the deep learning-based methods can be grouped by the ways to deal with spatial-temporal traffic data. One type of studies convert the spatial-temporal data into a 2-dimensional (2D) matrix and use long short-term memory (LSTM) recurrent neural network \cite{ma2015long}, bi-directional LSTM \cite{cui2016deep}, CNN \cite{ma2017learning}, GAN \cite{liang2018deep}, or a combination of multiple models \cite{yang2019cell}, to extract feature and forecast traffic states. However, a traffic network's spatial features cannot be completely represented by a 2D matrix. Thus, another type of methods \cite{yu2017spatiotemporal, ma2018forecasting} is proposed to convert the physical roadway networks as images according to roads' geospatial properties. Although the traffic network images demonstrate the true traffic network structure, those images contain too many noisy pixels and blank pixels without traffic state information. To analyze the traffic network in an efficient way, many studies consider the traffic network as a graph and predict traffic state by incorporating the graph convolutional network \cite{yu2017spatio, cui2018traffic, li2017diffusion}. 

\subsection{Deep learning-based traffic forecasting with missing values} 
Traffic forecasting performance will be highly influenced by the missing values. A bunch of data imputation methods has been developed to solve the missing values issues, including the probabilistic principal component analysis \cite{li2014traffic}, tensor decomposition-based methods \cite{ran2016tensor, Chen2019, chen2018spatial}, clustering approaches \cite{tang2015hybrid, ku2016clustering}. There are also some deep learning-based data imputation methods proposed in the most recent years, such as denoising stacked auto-encoder \cite{duan2016efficient} and generative adversarial imputation network \cite{yoon2018gain}. However, those deep learning-based methods requiring a large dataset covering a long period of time may not be feasible for the short-term traffic forecasting tasks.

To combine the data imputation and traffic forecasting together, a few RNN-based approaches, such as the LSTM-M \cite{tian2018lstm}, have been proposed based the GRU-D \cite{che2018recurrent} for processing multivariate time series with missing values. Even though these RNN-based methods can recurrently fill missing values in each time step and forecasting the future traffic state, they cannot capture spatial interactions between road links in the traffic network. Further, the RNN-based methods are uninterpretable like a black box. The models cannot tell the spatial relationship between neighboring links and the links' temporal dependencies between different time steps. 

To solve these problems, in this study, we consider the traffic state transition process as a graph Markov process and propose the graph Markov network for the traffic forecasting with missing values. The design of the graph Markov network inherently incorporates the spatial-temporal relationships among the links in the traffic network. The graph Markov network making use of the topological structure of the traffic network can achieve accurate prediction results with efficient training and testing process. The proposed graph Markov process and graph Markov network are introduced in detail in the following section.

\section{Problem Definition and Preliminary}

\subsection{Traffic Forecasting}

A traffic network normally consists of multiple roadway links. The traffic forecasting task targets to predict future traffic states of all (road) links or sensor stations in the traffic network based on historical traffic state data. The collected spatial-temporal traffic state data of a traffic network with $S$ links/sensor-stations can be characterized as a $T$-step sequence $[x_1,x_2,...,x_t,...,x_T] \in \mathbb{R}^{T \times S}$, in which $x_t \in \mathbb{R}^{S}$ demonstrates the traffic states of all $S$ links at the $t$-th step. The traffic state of the $s$-th link at time $t$ is represented by $x_t^s$. In this study, the superscript of a traffic state represents the spatial dimension and the subscript denotes the temporal dimension. The short-term traffic forecasting problem can be formulated as, based on $T$-step historical traffic state data, learning a function $F(\cdot)$ to generate the traffic state at next time step as follows:
\begin{equation} \label{eq:1}
    F([x_1,x_2...,x_T]) = [x_{T+1}]
\end{equation}

\subsection{Graph Representations of Traffic Network} 
Since the traffic network is composed of road links and intersections, it is intuitive to consider the traffic network as an undirected graph consisting of vertices and edges. The graph can be denoted as $\mathcal{G}=(\mathcal{V}, \mathcal{E}, \mathcal{A}, \mathcal{D})$ with a set of vertices $\mathcal{V}=\{v_1,...,v_D\}$ and a set of edges $\mathcal{E}$ between vertices. $\mathcal{A} \in \mathbb{R}^{S \times S}$ is a symmetric (typically sparse) adjacency matrix with binary elements, where $\mathcal{A}_{i,j}$ denotes the connectedness between nodes $v_i$ and $v_j$. The existence of an edge is represented through $\mathcal{A}_{i,j}=\mathcal{A}_{j,i}=1$, otherwise $\mathcal{A}_{i,j}=0$ ($\mathcal{A}_{i,i} = 0$). Based on $\mathcal{A}$, a diagonal graph degree matrix $\mathcal{D}\in \mathbb{R}^{S \times S}$ describing the number of edges attached to each vertex can be obtained by $\mathcal{D}_{i,i} = \sum_{j=1}^S \mathcal{A}_{i,j}$. 

The $\mathcal{A}$ can only indicate the relationship between different vertices. In some cases, the vertices' relationship with themselves also needs to be characterized. Thus, we define the self-connection adjacency matrix $\mathbf{A} = \mathcal{A} + I$, i.e. $\mathbf{A}_{i,i}=1$, which implies each vertex in the graph is self-connected. Here, $I \in \mathbb{R}^{S \times S}$ is an identity matrix. 

In addition, the connectedness of the graph vertices can also be encoded by the Laplacian matrix, which is essential for spectral graph analysis. The combinatorial Laplacian matrix is defined as $\mathcal{L} = \mathcal{D}-\mathcal{A}$ and the normalized definition is $\mathcal{L} = I - \mathcal{D}^{-1/2}\mathcal{A}\mathcal{D}^{-1/2}$. Since $\mathcal{L}$ is a symmetric positive semi-definite matrix, it can be diagonalized as $\mathcal{L}=U\Lambda U^T$ by its eigenvector matrix $U$ \cite{defferrard2016convolutional}, where $U=[u_0, u_1,..., u_{S-1}] \in \mathbb{R}^{S \times S}$ and $\Lambda = \operatorname{diag}(\lambda_0, \lambda_1,..., \lambda_{S-1})\in \mathbb{R}^{S \times S}$ is the corresponding diagonal eigenvalue matrix satisfying $\mathcal{L}u_i = \lambda_i u_i$.

In this study, under the traffic forecasting scenario, the attribute on vertex $v_s$ (road link $s$) at time $t$ is denoted as $x_t^s$. Given the graph representation of the traffic network, the Equation \ref{eq:1} can be extended as  
\begin{equation}
    F(\mathcal{G},[x_1,x_2...,x_T]) = [x_{T+1}]
\end{equation}

\subsection{Missing Values} 

Traffic state data can be collected by multi-types of traffic sensors or probe vehicles. When traffic sensors fail or no probe vehicles go through road links, the collected traffic state data may have missing values. We use a sequence of masking vectors $[m_1,m_2,...,m_T] \in \mathbb{R}^{T \times S}$, where $m_t \in \mathbb{R}^{S}$, to indicate the position of the missing values in traffic state sequence $[x_1,x_2,...,x_T]$. The masking vector can be obtained by
\begin{equation} \label{eq:3}
m_t^s = \left\{
\begin{aligned}
1 & \mbox{, if } x_t^s \mbox{ is observed} \\
0 & \mbox{, otherwise}
\end{aligned}
\right.
\end{equation}
where $x_t^s$ is the traffic state of $s$-th link at step $t$.

\subsection{Traffic Forecasting with Missing Values}

Missing values in traffic data can be handled by many existing data imputation methods. Most state-of-the-art data imputation methods, such as the Bayesian tensor decomposition approach \cite{Chen2019} and the Generative Adversarial Imputation Nets (GAIN) \cite{yoon2018gain}, need long-term historical data to capture complicated traffic patterns and fill missing values. However, in real-time environments, especially under the connected autonomous vehicle (CAV) and edge computing scenarios, it may not be possible to conduct data imputation on historical data and forecast future traffic states sequentially, because the volume of traffic state data is huge and the computing capability of devices is limited. In these cases, the traffic forecasting models should be able to handle missing values. Taking the missing values into consideration, we can formulate the traffic forecasting as follows
\begin{equation}
    F(\mathcal{G},[x_1,x_2...,x_T],[m_1,m_2...,m_T]) = [x_{T+1}]
\end{equation}

% \subsubsection{Relationship and difference between Data Imputation}

\section{Proposed Approach}
In this section, we first describe several properties of traffic states. Based on that, we propose a graph Markov process to characterize the variations of traffic states. Then, we introduce the proposed graph Markov Network for traffic forecasting with the capability of dealing with missing values.

\subsection{Properties}

A traffic network is a dynamic system and the states on all links keep varying resulted by the movements of vehicles in the system. Thus, we assume the traffic network's dynamic process satisfies the Markov property that the future state of the traffic network is conditional on the present state.

\textbf{Markov property:} The future state of the traffic network $x_{t+1}$ depends only upon the present state $x_t$, not on the sequence of states that preceded it. Taking $X_1, X_2, ... ,X_{t+1}$ as random variables with the Markov property and $x_1,x_2,...,x_{t+1}$ as the observed traffic states. The Markov process can be formulated in a conditional probability form as 
\begin{equation} \label{eq:5}
Pr(X_{t+1} = x_{t+1} | X_1 = x_1, X_2 = x_2, ..., X_t = x_t)=Pr(X_{t+1} = x_{t+1} | X_t = x_t)
\end{equation}
where $Pr(\cdot)$ demonstrates the probability. 

However, the transition matrix is temporal dependent, since at the different time of a day, the traffic state's transition pattern should be different. Based on Equation \ref{eq:5}, the transition process of traffic states can be formulated in the vector form as 
\begin{equation} \label{eq:6}
    x_{t+1} = P_t x_t
\end{equation}
where $P_t \in \mathbb{R}^{S\times S}$ is the transition matrix and ${(P_t)}_{i,j}$ measures how much influence $x_t^j$ has on forming the state $x_{t+1}^i$.

The transition process defined in Equation \ref{eq:6} does not take the time interval between $x_{t+1}$ and $x_t$ into consideration. We denote the time interval between two consecutive time steps of traffic states by $\Delta t$. If $\Delta t$ is small enough ($\Delta t \to 0$), the traffic network's dynamic process can be measured as a continuous process and the difference between consecutive traffic states are close to zero, i.e. $|x_{t+\Delta t} - x_t| \to 0$. However, a long time interval may result in more variations between the present and future traffic states, leading to a more complicated transition process. Since the traffic state data are normally processed into discrete data and the size of transition matrix $P_t$ is fixed, we consider that the longer the $\Delta t$ is, the lower capability of measuring the actual transition process $P_t$ has. Thus, we multiply a decay parameter $\gamma \in (0,1)$ in Equation \ref{eq:6} to represent the temporal impact on transition process as 
\begin{equation} \label{eq:7}
    x_{t+1} = \gamma P_t x_t
\end{equation}

% \subsubsection{Traffic Network's Spatial Influence on Traffic State}

The transition matrix can measure the contributions made by all roadway links on a specific link, which assumes that the state of a roadway link is influenced by all links in the traffic network. However, since vehicles in the traffic network traverse connected road links one by one and traffic states of connected links are transmitted by those vehicles, the traffic state of a link will only be affected by its neighboring links during a short period of time. 

\textbf{Graph localization property:} The traffic state of a specific link $s$ in a traffic network is mostly influenced by localized links, i.e. the link $s$ itself and its neighboring links, during a short period of time. The neighboring links refer to the links in the graph within a specific order of hops with respect to the link $s$. With the help of the graph's  topological structure, the localization property in the graph can be measured based the adjacency matrix in two ways: (1) The self-connection adjacency matrix $\mathbf{A}$, describing the connectedness of vertices, can inherently indicate the localization property of all vertices in the graph. Then, the impacts of localized links can be easily measured by a weighted self-connection adjacency matrix. (2) The other way is to conduct the spectral graph convolution operation on the traffic state $x_t$ to measure the localized impacts in the graph. The spectral graph convolution on $x_t$ can be defined as $U\Lambda_\theta U^T x_t$ \cite{defferrard2016convolutional}, where $U$ is the eigenvector matrix of the Laplacian matrix $\mathcal{L}$ and $\Lambda_\theta$ is a learnable diagonal weight matrix.

% Then, the $k$-hop localized relationship can be easily described by the $k$-th power of $\mathbf{A}$, i.e. $\mathbf{A}^k$. 

% The $\mathcal{L}$ can be written as an eigen decomposition as .
% Then, the impacts of localized links can be measured by a weighted self-connection adjacency matrix ($\mathbf{A} \odot W$), where $W$ is the weight matrix and $\odot$ is the Hadamard product operator.

\textbf{Graph Markov Process:} With the aforementioned two properties, we define the traffic state transition process as a graph Markov process (GMP). The graph Markov process can be formulated in a conditional probability form as
\begin{equation} 
Pr(X_{t+1} = x_{t+1}^u | X_t = x_t) = Pr(X_{t+1} = x_{t+1}^u | X_t = x_t^v, v \in \mathcal{N}(u))
\end{equation}
where the superscripts $u$ and $v$ are the indices of graph links (road links). The $\mathcal{N}(u)$ denotes a set of neighboring links of link $u$ and link $u$ itself. The properties of this graph Markov process is similar to the properties of the Markov random field \cite{rue2005gaussian} with temporal information. If we assume that road links are only influenced by their one-hop neighbors in the graph, based on Equation \ref{eq:7}, we can easily incorporate the graph localization properties into the traffic states' transition process by element-wise multiplying the transition matrix $P_t$ with the self-connection adjacency matrix $\mathbf{A}$. Then, the GMP can be formulated in the vector form as
\begin{equation} \label{eq:GMP}
    x_{t+1} = \gamma (\mathbf{A} \odot P_t) x_t
\end{equation}
where $\odot$ is the Hadamard (element-wise) product operator that $(\mathbf{A} \odot P_t)_{ij} = \mathbf{A}_{ij} \times {(P_t)}_{ij}$. 

The graph localization property can also be incorporated in the transition process by replacing the transition weight matrix $P_t$ with the spectral graph convolution operation. Then, we define the spectral version of the graph Markov process (SGMP) as
\begin{equation}
    x_{t+1} = \gamma U \Lambda_{\theta_t} U^T x_t
\end{equation}
where $\Lambda_{\theta_t} \in \mathbb{R}^{S \times S}$ is a diagonal weight matrix.

% It should be noted that $\mathbf{A}^k$ normally has elements larger than one. 

% if we do not incorporate the graph, it is a combinatorial fully connected NN

% Unobserved 
% uncertanty is removed.

\begin{figure*}[t!]
  \includegraphics[width =  \textwidth]{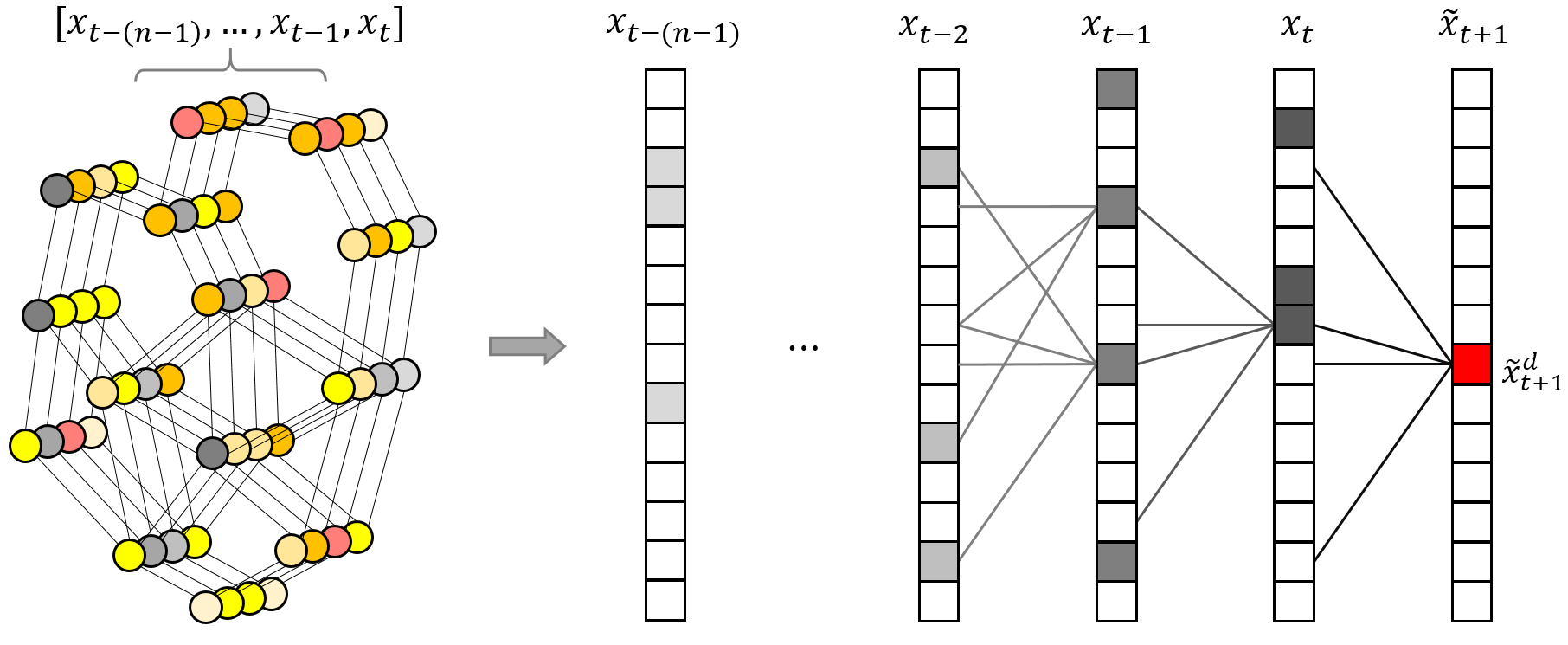}
  \caption{Graph Markov process. The gray-colored nodes in the left sub-figure demonstrating the nodes with missing values. The traffic states are represented by vectors on the right side. A future state (in red color) and the missing states (in gray color) can be inferred from their neighbors in the previous step.}
  \label{fig:graphMarkovProcess}
\end{figure*}

\subsection{Handling Missing Values in the Graph Markov Process}
In this section, we theoretically introduce how to deal with the missing values in the graph Markov process. 

As we assume the traffic state transition process follows the graph Markov process, the future traffic state can be inferred by the present state. If there are missing values in the present state, we can infer the missing values from previous states. We consider $x_t$ is the observed traffic state at time $t$ and a mask vector $m_t$ can be acquired according to Equation \ref{eq:3}. We denote the completed state by $\tilde{x}_t$, in which all missing values are filled based on historical data. Hence, the completed state consists of two parts, including the observed state values and the inferred state values, as follows:
\begin{equation} \label{eq:completedState}
    \tilde{x}_t = x_t \odot m_t + \tilde{x}_t \odot (1-m_t)
\end{equation}
where $\tilde{x}_t \odot (1-m_t)$ is the inferred part. Since $x_t \odot m_t = x_t$, Equation \ref{eq:completedState} can be written as 
\begin{equation} \label{eq:completedState_2}
    \tilde{x}_t = x_t + \tilde{x}_t \odot (1-m_t)
\end{equation}

Since the transition of completed states follows the the graph Markov process, the GMP and SGMP with respect to the completed state can be described as 
$\tilde{x}_{t+1} = \gamma (\mathbf{A} \odot P_t) \tilde{x}_t$ and $\tilde{x}_{t+1} = \gamma U \Lambda_{\theta_t} U^T \tilde{x}_t$, respectively. In this section, for simplicity, we denote the (spectral) graph Markov transition matrix by $H_t$, i.e. $H_t = \mathbf{A} \odot P_t$ or $H_t = U \Lambda_{\theta_t} U^T$. Hence, the transition process of completed states can be represented as
\begin{equation}
    \tilde{x}_{t+1} = \gamma H_t \tilde{x}_t 
\end{equation}
Applying Equation \ref{eq:completedState_2}, the transition process becomes
\begin{equation} \label{eq:GMP_1_step}
     \tilde{x}_{t+1} = \gamma H_t (x_t + \tilde{x}_t \odot (1-m_t)) 
\end{equation}

If we iteratively apply the completed state $\tilde{x}_t$, i.e. $\tilde{x}_t = \gamma H_{t-1} (x_{t-1} + \tilde{x}_{t-1} \odot (1-m_{t-1}))$, into Equation \ref{eq:GMP_1_step} itself, we have
\begin{align}
    \tilde{x}_{t+1} = &\; \gamma H_t (x_t + \gamma H_{t-1} (x_{t-1} + \tilde{x}_{t-1} \odot (1-m_{t-1})) \odot (1-m_t)) \notag \\
    = &\; \gamma H_t x_t + \gamma^2 H_t H_{t-1} (x_{t-1} \odot (1-m_t)) + \gamma^2 H_t H_{t-1} (\tilde{x}_{t-1} \odot (1-m_{t-1}) \odot (1-m_t))
\end{align}
After iteratively applying $n$ steps of previous states from $x_{t-(n-1)}$ to $x_t$, $\tilde{x}_{t+1}$ can be described as
\begin{align} \label{eq:GMP_n_step}
    \tilde{x}_{t+1} = \; & \gamma H_t x_t \notag \\
        + \;& \gamma^2 H_t H_{t-1} (x_{t-1} \odot (1-m_t))  \notag \\
        + \;& \gamma^3 H_t H_{t-1} H_{t-2} (x_{t-2}\odot (1-m_{t-1}) \odot (1-m_t)) \notag + \cdots \notag \\ 
        % &\; + \gamma^4 H_t H_{t-1} H_{t-2} H_{t-3}(\tilde{x}_{t-3} \odot (1-m_{t-2}) \odot (1-m_{t-1}) \odot (1-m_t)) + \cdots \notag \\
        + \;& \gamma^{n} H_t \cdots H_{t-(n-1)} (x_{t-(n-1)} \odot (1-m_{t-(n-2)}) \odot \cdots \odot (1-m_t)) \notag \\
        + \; & \gamma^{n} H_t \cdots H_{t-(n-1)} (\tilde{x}_{t-(n-1)} \odot (1-m_{t-(n-1)}) \odot \cdots \odot (1-m_t))
\end{align}
The $n$ steps of historical steps of states can be written in a summation form as 
\begin{align} \label{eq:GMP_full}
    \tilde{x}_{t+1} = &\; \sum_{i=0}^{n-1} \gamma^{i+1} (\prod_{j=0}^{i} H_{t-j}) (x_{t-i} \odot \bigodot_{j=0}^{i-1} (1-m_{t-j})) \notag \\ 
         + &\; \gamma^{n} H_t \cdots H_{t-(n-1)} (\tilde{x}_{t-(n-1)} \odot (1-m_{t-(n-1)}) \odot \cdots \odot (1-m_t))
\end{align}
where $\sum$, $\prod$, and $\bigodot$ are the summation, matrix product, and Hadamard product operators, respectively. For simplicity, we denote the term with the $\tilde{x}_{t-n}$ in Equation \ref{eq:GMP_full} as $\mathcal{O}(\tilde{x}_{t-n})$, and the GMP of the complected states can be represented by
\begin{equation}\label{eq:GMP_def}
    \tilde{x}_{t+1} = \sum_{i=0}^{n-1} \gamma^{i+1} (\prod_{j=0}^{i} H_{t-j}) (x_{t-i} \odot \bigodot_{j=0}^{i-1} (1-m_{t-j})) + \mathcal{O}(\tilde{x}_{t-(n-1)}) 
\end{equation}

In $\mathcal{O}(\tilde{x}_{t-(n-1)})$, when $n \rightarrow \infty$, since $\gamma \in (0,1)$, $\gamma^{n+1}$  $\rightarrow 0$. In addition, the product of masking vectors in $\mathcal{O}(\tilde{x}_{t-(n-1)})$ will also approach to zero, i.e. $\bigodot_{j=0}^{n-1} (1-m_{t-j}) \rightarrow 0$. If the missing rate of the traffic state values is 10\% and the missing pattern is random, the probability of each element of $\bigodot_{j=0}^{n-1} (1-m_{t-j})$ being zero is $10^{n}$. Thus, when $n$ is large enough, we consider the $\mathcal{O}(\tilde{x}_{t-(n-1)})$ is a negligibly term.
% and denote the future complected state $\tilde{x}_{t+1}$ as

% \begin{equation} \label{eq:GMP_def}
% \tilde{x}_{t+1} = \sum_{i=0}^{n-1} \gamma^{i+1} (\prod_{j=0}^{i} H_{t-j}) (x_{t-i} \odot \bigodot_{j=0}^{i-1} (1-m_{t-j}))
% \end{equation}

Figure \ref{fig:graphMarkovProcess} demonstrates the graph Markov process for inferring the future state. The traffic network graphs with attribute-missed nodes (in gray color) is convert into traffic state vectors. The inference of $\tilde{x}_{t+1}^d$ is based on historical traffic states by back-propagating to the $t-(n-1)$ step.

% To make simplify the neural network, we define the 
% \begin{equation}
% \tilde{x}_{t+1} = \sum_{i=0}^{n} \gamma^i H_i x_{t-i} \odot \bigodot_{j=0}^i (1-m_{t-j})
% \end{equation}
% where $H_i = \prod_{j=0}^{i} A_{t-j}$

\subsection{Graph Markov Network}

\begin{figure*}[t!]
  \includegraphics[width =  \textwidth]{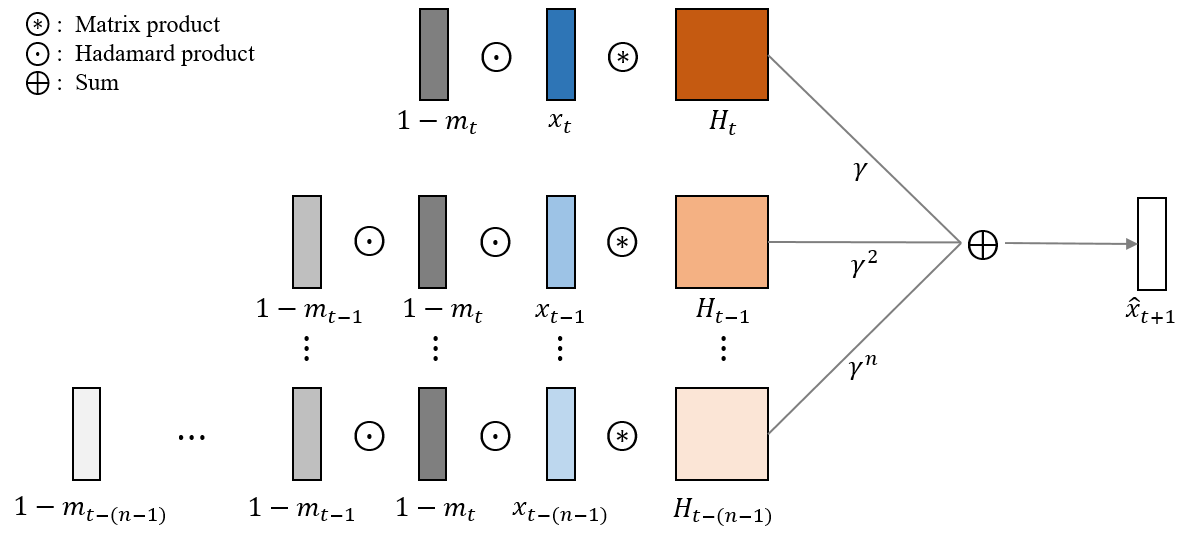}
  \caption{Structure of the proposed graph Markov network. Here, $H_{t-j} = \mathbf{A}^j\odot W_j$. As for the spectral version of GMN, $H_{t-j}=U \Uplambda_j U^T$.}
  \label{fig:graphMarkovNetwork}
\end{figure*}

In this section, we propose a \textbf{Graph Markov Network} (GMN) for traffic prediction with the capability of handling missing values in historical data. Suppose the historical traffic data consists of $n$ steps of traffic states $\{x_{t-(n-1)}, ..., x_t\}$. Correspondingly, we can acquire $n$ masking vectors $\{m_{t-(n-1)}, ..., m_t\}$. The traffic network's topological structure can be represented by the adjacency matrix. 

The GMN is designed based on the proposed GMP described in the previous section. Since we consider the term $\mathcal{O}(\tilde{x}_{t-(n-1)})$ described in Equation \ref{eq:GMP_def} is small enough, the $\mathcal{O}(\tilde{x}_{t-(n-1)})$ is omitted in the proposed GMN for simplicity. 

As described in Equation \ref{eq:GMP_def}, the graph Markov process contains $n$ transition weight matrices and the product of the these matrices $(\prod_{j=0}^{i} H_{t-j}) = (\prod_{j=0}^{i} \mathbf{A}^j \odot P_{t-j})$ measures the contribution of $x_{t-i}$ for generating the $\tilde{x}_t$. To reduce matrix product operations and at the same time keep the learning capability in the GMP, we simplify the $(\prod_{j=0}^{i} \mathbf{A}^j \odot P_{t-j})$ by $(\mathbf{A}^{i+1} \odot W_{i+1})$, where $W_{i+1} \in \mathbb{R}^{S \times S}$ is a weight matrix. In this way, $(\mathbf{A}^{i+1} \odot W_{i+1})$ can directly measure the contribution of $x_{t-i}$ for generating the $\tilde{x}_t$ and skip the intermediate state transition processes. Further, the GMP still has $n$ weight matrices ($\{\mathbf{A}^1 \odot W_1, ..., \mathbf{A}^n \odot W_n\}$), and thus, the learning capability in terms of the size of parameters does not change. The benefits of the simplification is that the GMP can reduce $\frac{n(n-1)}{2}$ times of multiplication between two $S\times S$ matrices in total. 

Based on the GMP and the aforementioned simplification, we propose the \textbf{\textit{graph Markov network}} for traffic forecasting with the capability of handling missing values as

\begin{equation} \label{eq:GMN}
\hat{x}_{t+1} = \sum_{i=0}^{n-1} \gamma^{i+1} (\mathbf{A}^{i+1} \odot W_{i+1}) (x_{t-i} \odot \bigodot_{j=0}^{i-1} (1-m_{t-j}))
\end{equation}
where $\hat{x}_{t+1}$ is the predicted traffic state for the future time step $t+1$ and $\{W_1,...,W_n\}$ are the model's weight matrices that can be learned and updated during the training process. 
% The GMN can also be written as an expanded version that
% \begin{align} \label{eq:GMN_expansion}
%     \hat{x}_{t+1} = &\; \gamma \mathbf{A}W_1 x_t + \gamma^2 \mathbf{A}^2 W_2 (x_{t-1} \odot (1-m_t)) + \cdots \notag \\ 
%          + &\; \gamma^{n} \mathbf{A}^n W_n (x_{t-(n-1)} \odot (1-m_{t-(n-2)}) \odot \cdots \odot (1-m_t)) 
% \end{align}
% \gamma^3 \mathbf{A}^3 W_3 (x_{t-2}\odot (1-m_{t-1}) \odot (1-m_t))

As for the spectral version of the graph Markov process, the product of the transition weight matrices $(\prod_{j=0}^{i} H_{t-j})$ can also be simplified. Because of the orthogonality of the eigenvectors of $\mathcal{L}$ \cite{wang2015orthogonal}, $U^T = U^{-1}$, and thus, $(\prod_{j=0}^{i} H_{t-j}) = (\prod_{j=0}^{i} U \Lambda_{\theta_{t-j}} U^T) = U (\prod_{j=0}^{i} \Lambda_{\theta_{t-j}}) U^T$. We further simplify the product of the transition weight matrices by replacing the $\prod_{j=0}^{i} \Lambda_{\theta_{t-j}}$ with a diagonal weight matrix $\Uplambda_{i+1}$. Similar to the simplification of GMP, this simplification process will not reduce the learning capability of the SGMP because the amount of the weight parameters does not change. In this way, the SGMP can reduce at least $\frac{n(n-1)}{2}$ times of vector multiplication in total. Then, the spectral version of the graph Markov network (SGMN) can be defined as 
\begin{equation} \label{eq:SGMN}
\hat{x}_{t+1} = \sum_{i=0}^{n-1} \gamma^{i+1} (U \Uplambda_{i+1} U^T) (x_{t-i} \odot \bigodot_{j=0}^{i-1} (1-m_{t-j}))
\end{equation}
where $\{\Uplambda_1,...,\Uplambda_n\}$ are the diagonal weight matrices that can be learned and updated during the training process.

The structure of GMN for predicting traffic state $x_{t+1}$ is demonstrated in Figure \ref{fig:graphMarkovNetwork}. The spectral version of GMN has the same model structure that it only need to replace the $\mathbf{A}^j\odot W_j$ with the $U \Uplambda_j U^T$, as shown Figure \ref{fig:graphMarkovNetwork}. During the training process, the loss can be calculated by measuring the difference between the predicted value $\hat{y}=\hat{x}_{t+1}$ and the label $y = x_{t+1}$.

% \subsubsection{proof and graph convolutional version}

% \subsubsection{relationship to RNN}

% \subsubsection{relationship to Dropout}

% \subsubsection{relationship to fully connected NN}

% \subsection{Comparison between GMN and RNN}

% \pagebreak

\section{Experimental Results}
In this section, we compared the proposed approach with state-of-the-art traffic forecasting models with the capability of handling missing values. The datasets, hyper-parameters, software, and hardware used in the experiments are introduced. 

\subsection{Datasets}
In this study, we conduct experiments on three real-world network-wide traffic state datasets. The topological structures of the traffic networks are also used in the experiments.

\subsubsection{PEMS-BAY}
This dataset named as PEMS-BAY is collected by California Transportation Agencies (CalTrans) Performance Measurement System (PeMS). This dataset contains the speed information of 325 sensors in the Bay Area lasting for six months ranging from Jan 1st, 2017 to Jun 30th, 2017. The interval of time steps is 5-minutes. The total number of observed traffic data points is 16,941,600. The adjacency matrix of this dataset is defined according to \cite{li2018dcrnn_traffic}. The dataset is published by \cite{li2018dcrnn_traffic} on the Github (\url{https://github.com/liyaguang/DCRNN}).

\subsubsection{METR-LA}
This dataset is collected from loop detectors on the freeway of Los Angeles County \cite{jagadish2014big}. This dataset contains the speed information of 207 sensors lasting for 4 months ranging from Mar 1st, 2012 to Jun 30th, 2012. The interval of time steps is 5-minutes. The total number of observed traffic data points is 6,519,002. Similar to the PEMS-BAY dataset, the adjacency matrix of this dataset is defined according to \cite{li2018dcrnn_traffic}, and the dataset is published on the Github (\url{https://github.com/liyaguang/DCRNN}).

\subsubsection{INRIX-SEA}
This dataset is collected by the INRIX company from multiple data sources, including GPS probes, road sensors, and cell phone data. This dataset contains the speed information of the traffic network in the Seattle downtown area consisting of 725 road segments. The traffic network covers both freeways and urban roadways. The  dataset covers a one-year period from Jan 1st, 2012 to Dec 31st, 2012. The interval of time steps is 5-minutes. The total number of observed traffic data points is 76,212,000. This dataset is provided by Washington Department of Transportation and has been used in \cite{cui2018traffic}. Due to privacy policies, this dataset is not published.

% \subsubsection{LOOP-SEA}
% This dataset is collected by Washington State Department of Transportation from loop detectors on the freeway of the Seattle metropolitan area and processed by pre-processed by the University of Washington \cite{cui2016deep}. This dataset contains the speed information of 323 sensor lasting for one year from Jan 1st 2015 to Dec 31st 2015. The interval of time steps is 5-minutes. The total number of observed traffic data points is 33,953,760. The dataset is published on Zenodo \cite{yinhai_wang_2019_3258904} and Github (\url{https://github.com/zhiyongc/Seattle-Loop-Data})

\subsection{Missing Values and Data formatting}

The dataset forms as a spatial-temporal matrix, whose spatial dimension size is the number of sensors and temporal dimension size is the number of time steps. In the experiments, the dataset is split into a training set, a validation set, and a testing set, with a size ratio of 6:2:2. In the training and testing process, the speed values of the dataset are normalized within a range of [0,1].

% the sample sizes of the PEMS-BAY, METR-LA, and LOOP-SEA datasets are 52,119, 35,127, and 105,111, respectively. 

The PEMS-BAY and METR-LA datasets originally have missing values and their percentages of missing values are 0.003\% and 8.11\%, respectively. There are no missing values in the INRIX-SEA dataset. To test the model's capability of handling missing values with different missing rates, we randomly set a portion of speed values in the input sequences as zeros according to a specific missing rate and generate the masking vectors accordingly. In this study, based on each of the three datasets, three sub-datasets with missing rates of 10\%, 20\%, and 40\%, respectively, are generated.

\subsection{Hardware and Software Environments}
The experiments are conducted on a computer with 128GB memory, a Intel Core i9-7900X CPU, and a NVIDIA GTX 1080 Ti GPU. The proposed approach and all neural network-based baseline models are implemented based on Pytorch 1.0.1 using the Python language 3.6.8.

\subsection{Baseline Models}
\begin{itemize}
  \item GRU \cite{cho2014learning}: GRU referring to gated recurrent units is a type of RNN. GRU can be considered as a simplified LSTM.
  \item GRU-I : GRU-I is designed based on GRU. Since GRU is a type of a RNN with the recurrent structure, the predicted values from a previous step $\hat{x}_t$ can be used to infer the missing values in the next step. The completed traffic states with all missing values filled can be represented by $\tilde{x}_{t+1} = x_t + \hat{x}_t \odot (1-m_t)$.
  \item GRU-D \cite{che2018recurrent}: GRU-D is a neural network structure that is designed based on GRU for multivariate time series prediction. It can capture long-term temporal dependencies in time series. GRU-D incorporates the masking information and missing values' time interval as input such that it can utilize the missing patterns.
  \item LSTM \cite{hochreiter1997long}: LSTM is a powerful variant of RNN, which can overcome the gradients exploding or vanishing problem. It is suitable for being a model's basic structure for traffic forecasting.
  \item LSTM-I : LSTM-I is designed based on LSTM. The missing value inferring process of LSTM-I is similar to that of GRU-I.
  \item LSTM-M \cite{tian2018lstm}: LSTM-M is a neural network structure designed based on LSTM for traffic prediction with missing data. It employs multi-scale temporal smoothing methods to infer lost data.
%   \item Vector autoregression (VAR)-based neural network (VARNN) \cite{zhang2016comparative}: 
\end{itemize} 

\subsection{Model Parameters}
The batch size of the tested data is set as 64. The number of steps of historical data incorporated in the GMN model will have an influence on the prediction performance. Hence,the GMNs with 6-steps, 8-steps, and 10-steps of historical data are tested in the experiments, i.e. the $n$ in Equations \ref{eq:GMN} and \ref{eq:SGMN} are set as 6, 8, and 10. In the following sections, we denoted these GMN models as GMN-6, GMN-8, and GMN-10, respectively. The corresponding SGMN models with different steps are denoted as SGMN-6, SGMN-8,and SGMN-10, respectively. The decay parameter $\gamma$ is set as 0.9 in the experiments. For the RNN-based baseline models, including GRU, GRU-I, GRU-D, LSTM, LSTM-I, and LSTM-M, their input sequences all have 10 time steps.

\subsection{Training and Hyper-Parameters}
In the training process, the mean square error (MSE) between the label $y_t$ and the predicted value $\hat{y}_t$, i.e. $\frac{1}{n}\sum_{i=1}^N(y_{i} - \hat{y}_i)^{2}$ is used as the loss function, where $N$ is the sample size. The Adam \cite{kingma2014adam} optimization method is adopted for both GMN models and baseline models to update parameters, as Adam is also used in  \cite{che2018recurrent, tian2018lstm}. 
Early stopping mechanism is used to avoid over-fitting. If there is no improvement in 5 consecutive epochs, the training will be stopped. The minimum improvement in MSE is set as 0.00001. We also design a learning rate decay mechanism for the training process to speed up the models' convergence. The initial learning rate of all models is set as $10^{-3}$, which is identical to the learning rate in \cite{tian2018lstm}. If there is no improvement in 4 consecutive epochs, the learning rate is reduced an order of magnitude until it reaches $10^{-5}$.

\subsection{Evaluation Metric}
The prediction accuracy of all tested models are evaluated by three metrics, including mean absolute error (MAE), mean absolute percentage error (MAPE), and root mean square error (RMSE).
\begin{equation}
    MAE = \frac{1}{N}\sum_{i=1}^N |y_i - \hat{y}_i|
\end{equation}

\begin{equation}
    MAPE = \frac{1}{N}\sum_{i=1}^N|\frac{y_i - \hat{y}_i}{y_i}|
\end{equation}

\begin{equation}
    RMSE = (\frac{1}{N} \sum_{i=1}^N {|y_i - \hat{y}_i|}^2 )^{\frac{1}{2}}
\end{equation}

\subsection{Experimental Results}

\begin{table}[!htb]
\caption {Prediction performance on PEMS-BAY dataset}
\label{tab:result_pems_bay}
\resizebox{\textwidth}{!}{%
\begin{tabular}{|l||c|c|c|c|c|c|c|c|c|}
\hline
\multicolumn{10}{|c|}{PEMS-BAY}                                                                                                                      \\ \hline
\multirow{2}{*}{Model} & \multicolumn{3}{c|}{Missing Rate = 10\%} & \multicolumn{3}{c|}{Missing Rate = 20\%} & \multicolumn{3}{c|}{Missing Rate = 40\%} \\ \cline{2-10} 
                       & MAE         & MAPE(\%)         & RMSE  
                       & MAE         & MAPE(\%)         & RMSE  
                       & MAE         & MAPE(\%)         & RMSE       \\ \hline\hline
GRU                    & 1.608       & 3.133        & 2.608      & 1.787       & 3.522        & 2.911      & 2.052       & 4.095       & 3.320        \\ \hline
GRU-I                  & 1.108       & 2.133        & \hl{1.831} & 1.185       & 2.296        & \hl{1.968} & 1.385       & 2.729       & \hl{2.327}       \\ \hline
GRU-D                  & 5.320       & 13.584       & 9.163      & 5.347       & 13.609       & 9.160      & 5.387       & 13.67       & 9.180        \\ \Xhline{2\arrayrulewidth}
LSTM                   & 2.368       & 4.809        & 3.952      & 2.457       & 5.098        & 4.258      & 2.428       & 5.117       & 4.181       \\ \hline
LSTM-I                 & 2.218       & 4.001        & 7.472      & 2.373       & 4.278        & 7.742      & 2.058       & 3.989       & 5.863       \\ \hline
LSTM-D                 & 1.198       & 2.351        & 1.968      & 1.236       & 2.437        & 2.055      & 1.472       & 2.904       & 3.111       \\ \Xhline{2\arrayrulewidth}
GMN-6                  & 1.084       & 2.077        & 2.565      & 1.101       & 2.145        & 2.029      & 1.819       & 3.634       & 3.543       \\ \hline
GMN-8                  & 1.089       & 2.086        & 2.611      & 1.196       & 2.297        & 2.827      & 1.376       & 2.730       & 2.678       \\ \hline
GMN-10                 & 1.089       & 2.086        & 2.614      & 1.202       & 2.308        & 2.864      & 1.327       & 2.615       & 2.470        \\ \Xhline{2\arrayrulewidth}
SGMN-6                & 1.009       & 1.930        & 1.877      & 1.064       & 2.048        & 2.067      & 1.930       & 3.671       & 4.375       \\ \hline
SGMN-8                & 1.008       & 1.929        & 1.875      & 1.062       & 2.043        & 2.024      & 1.291       & 2.517       & 2.867       \\ \hline
SGMN-10               & \hl{1.007}  & \hl{1.927}   & 1.874      & \hl{1.058}  & \hl{2.037}   & 2.018      & \hl{1.207}  & \hl{2.362}  & 2.473       \\ \hline
\end{tabular}
}
\end{table}

\begin{table}[!htb]
\caption {Prediction performance on METR-LA dataset}
\label{tab:result_metr_la}
\resizebox{\textwidth}{!}{%
\begin{tabular}{|l||c|c|c|c|c|c|c|c|c|}
\hline
\multicolumn{10}{|c|}{METR-LA}                                                                                                                       \\ \hline
\multirow{2}{*}{Model} & \multicolumn{3}{c|}{Missing Rate = 10\%} & \multicolumn{3}{c|}{Missing Rate = 20\%} & \multicolumn{3}{c|}{Missing Rate = 40\%} \\ \cline{2-10} 
                       & MAE         & MAPE(\%)        & RMSE   
                       & MAE         & MAPE(\%)        & RMSE   
                       & MAE         & MAPE(\%)        & RMSE        \\ \hline\hline
GRU                    & 3.427       & 7.971       & 5.923       & 3.667       & 8.611       & 6.249       & 4.037       & 9.622       & 6.744       \\ \hline
GRU-I                  & 3.322       & 7.625       & 5.543       & 3.402       & 7.846       & 5.642       & \hl{3.389}  & 7.917       & 5.903       \\ \hline
GRU-D                  & 9.912       & 25.28       & 12.195      & 9.904       & 25.302      & 12.193      & 10.022      & 25.444      & 12.269      \\ \Xhline{2\arrayrulewidth}
LSTM                   & 3.477       & 8.050       & 6.015       & 3.652       & 8.559       & 6.263       & 3.899       & 9.300       & 6.663       \\ \hline
LSTM-I                 & 3.180       & 7.228       & 5.363       & \hl{3.267}  & 7.417       & 5.653       & 3.393       & \hl{7.826}  & 5.879       \\ \hline
LSTM-D                 & 3.253       & 7.374       & 5.540       & 3.368       & 7.666       & 5.717       & 3.410       & 7.837       & \hl{5.812}  \\ \Xhline{2\arrayrulewidth}
GMN-6                  & 3.384       & 7.300       & 5.624       & 3.477       & 7.488       & 5.583       & 3.913       & 8.518       & 6.362       \\ \hline
GMN-8                  & 3.565       & 7.684       & 6.126       & 3.653       & 7.852       & 6.001       & 3.864       & 8.365       & 6.083       \\ \hline
GMN-10                 & 3.708       & 7.969       & 6.512       & 3.792       & 8.131       & 6.411       & 3.961       & 8.518       & 6.216       \\ \Xhline{2\arrayrulewidth}
SGMN-6                & \hl{3.145}  & \hl{6.836}  & 5.331       & 3.333       & 7.232       & 5.578       & 3.952       & 8.593       & 6.894       \\ \hline
SGMN-8                & 3.174       & 6.889       & 5.362       & 3.318       & 7.203       & 5.552       & 3.699       & 8.053       & 6.186       \\ \hline
SGMN-10               & 3.152       & 6.852       & \hl{5.321}  & 3.310       & \hl{7.187}  & \hl{5.525}  & 3.680       & 8.005       & 6.079       \\ \hline

\end{tabular}
}
\end{table}

\begin{table}[!htb]
\caption {Prediction performance on INRIX-SEA dataset}
\label{tab:result_inrix_sea}
\resizebox{\textwidth}{!}{%
\begin{tabular}{|l||c|c|c|c|c|c|c|c|c|}
\hline
\multicolumn{10}{|c|}{INRIX-SEA}                                                                                                                     \\ \hline
\multirow{2}{*}{Model} & \multicolumn{3}{c|}{Missing Rate = 10\%} & \multicolumn{3}{c|}{Missing Rate = 20\%} & \multicolumn{3}{c|}{Missing Rate = 40\%} \\ \cline{2-10} 
                       & MAE         & MAPE(\%)         & RMSE  
                       & MAE         & MAPE(\%)         & RMSE  
                       & MAE         & MAPE(\%)         & RMSE       \\ \hline\hline
GRU                    & 1.097       & 3.964        & 2.158      & 1.146       & 4.143        & 2.257      & 1.256       & 4.530        & 2.443      \\ \hline
GRU-I                  & 0.888       & 3.220        & \hl{1.850} & 0.939       & 3.419        & \hl{1.920} & 1.057       & 3.889        & \hl{2.086} \\ \hline
GRU-D                  & 3.039       & 11.597       & 5.408      & 2.947       & 11.399       & 5.160      & 2.873       & 11.136       & 5.008      \\ \Xhline{2\arrayrulewidth}
LSTM                   & 1.256       & 4.451        & 2.446      & 1.450       & 5.364        & 2.956      & 1.433       & 5.260        & 2.902      \\ \hline
LSTM-I                 & 0.945       & 3.363        & 2.400      & 0.910       & 3.255        & 2.094      & 1.592       & 5.155        & 5.156      \\ \hline
LSTM-D                 & 1.096       & 4.357        & 2.633      & 1.001       & 3.787        & 2.264      & 0.986       & 3.584        & 2.098      \\ \Xhline{2\arrayrulewidth}
GMN-6                  & 2.354       & 8.541        & 4.832      & 2.704       & 9.588        & 5.545      & 3.063       & 10.700       & 5.960      \\ \hline
GMN-8                  & 2.356       & 8.547        & 4.835      & 2.712       & 9.613        & 5.559      & 2.938       & 10.277       & 5.803      \\ \hline
GMN-10                 & 2.355       & 8.545        & 4.835      & 2.713       & 9.618        & 5.561      & 2.923       & 10.224       & 5.778      \\ \Xhline{2\arrayrulewidth}
SGMN-6                & \hl{0.768}  & 2.715        & 1.922      & 0.829       & 2.940        & 2.038      & 1.355       & 4.949        & 2.983      \\ \hline
SGMN-8                & \hl{0.768}  & \hl{2.713}   & 1.921      & \hl{0.826}  & \hl{2.929}   & 2.026      & 1.024       & 3.679        & 2.399      \\ \hline
SGMN-10               & \hl{0.768}  & 2.716        & 1.921      & 0.827       & 2.934        & 2.026      & \hl{0.973}  & \hl{3.485}   & 2.283      \\ \hline
\end{tabular}
}
\end{table}

\subsubsection{Comparing with baseline models}
The prediction results tested on the PEMS-BAY, METR-LA, and INRIX-SEA datasets with respect to different missing rates are displayed in Table \ref{tab:result_pems_bay}, Table \ref{tab:result_metr_la}, and Table \ref{tab:result_inrix_sea}, respectively. Overall, the SGMN models are superior to other baseline models. The GMN models also perform well, especially on the PEMS-BAY dataset. However, the prediction performance of GMN models decreases faster than that of SGMN models along with the increase of the missing rate. Among the baseline models, the GRU-I model performs well that it achieve smaller RMSEs on the PEMS-BAY and INRIX-SEA datasets. 

As shown in Table \ref{tab:result_pems_bay}, the SGMN-10 achieves the smallest MAEs and MAPEs for all the three missing rates on the PEMS-BAY dataset. However, the RMSEs of GRU-I are the smallest ones for all missing rates. That means the GRU-I's prediction results tend to have less large errors. The test results on the INRIX-SEA dataset, as shown in Table \ref{tab:result_inrix_sea}, have the similar situation that SGMN models perform better in terms of the MAE and MAPE metrics and GRU-I achieves better RMSE results. As for the results tested on the METR-LA dataset, shown in Table \ref{tab:result_metr_la}, the SGMN models outperform other models when the missing rates are 10\% and 20\%. When the missing rate increases to 40\%, the GRU-I, LSTM-I, and LSTM-D models achieve better prediction performance in terms of all the three metrics. It should be noted that 8.11\% of the values in the METR-LA dataset are originally missing. The actual missing rates of the METR-LA dataset are higher than the rates we set in the experiments. Hence, the SGMN models can achieve superior prediction results, especially when the missing rate is not very high.

\subsubsection{Analysis on Training Time}

\begin{figure}[!htb]
    \centering
    \includegraphics[width=0.6\textwidth]{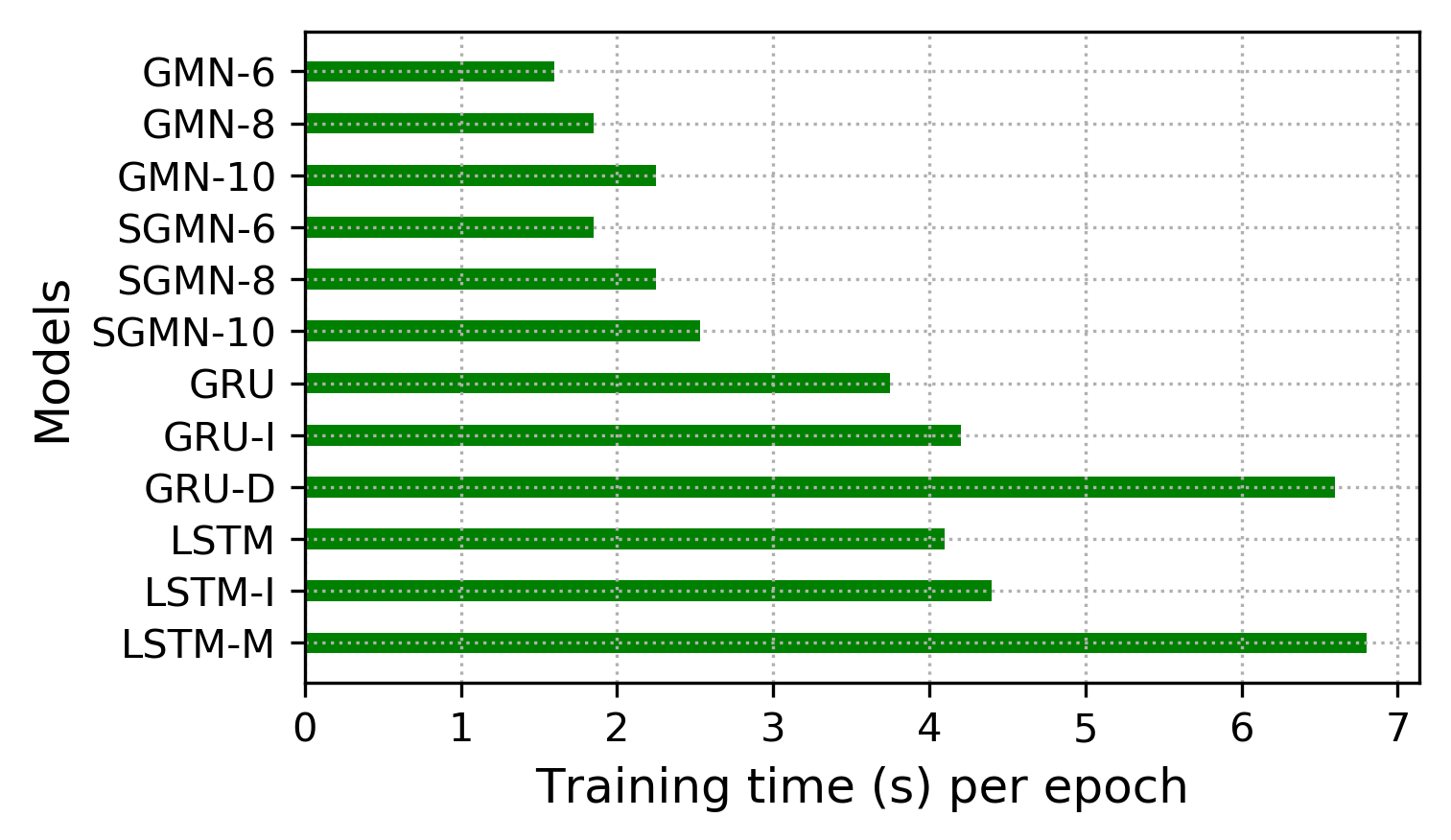}
    \caption{Training time of the compared models.}
    \label{fig:training_time}
\end{figure}

In this section, we analyze the training time of the proposed models and baseline models. Figure \ref{fig:training_time} shows the training time per epoch of the compared models tested on the PEMS-BAY datasets. The training times tested on different datasets have the same patterns. Since the GMN models have less matrix product operations than the SGMN models, GMN models take slightly less time per epoch than other models. The GMN and SGMN models apparently cost less running time than the baseline models because they get rid of the recurrent structure. The training times of GMN and SGMN increase when they incorporate more historical steps. The GRU and LSTM have similar training time per epoch. Since the GRU-I and LSTM-I have an imputation operation, their training times take a little bit more. In addition, since the GRU-D and LSTM-M both take more types of data as the input, their training time is much more than GRU and LSTM. 

\subsubsection{Analysis on Decay Rates of GMN and SGMN}

\begin{figure}[!htb]
     \centering
     \begin{subfigure}[b]{0.32\textwidth}
         \centering
         \includegraphics[width=\textwidth]{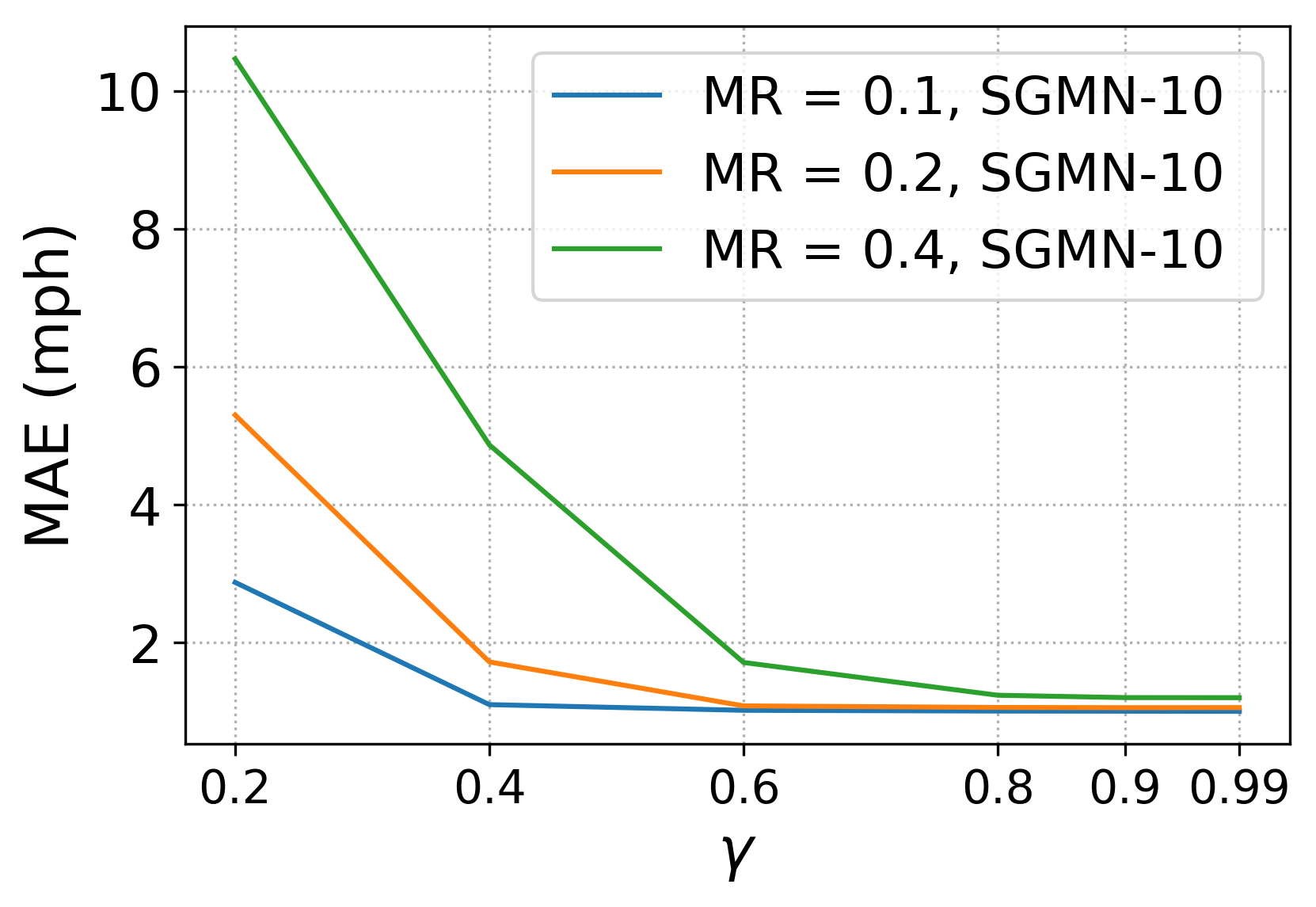}
         \caption{SGMN-10 on PEMS-BAY}
         \label{fig:gamma_sgmn_pems}
     \end{subfigure}
     \hfill
     \begin{subfigure}[b]{0.32\textwidth}
         \centering
         \includegraphics[width=\textwidth]{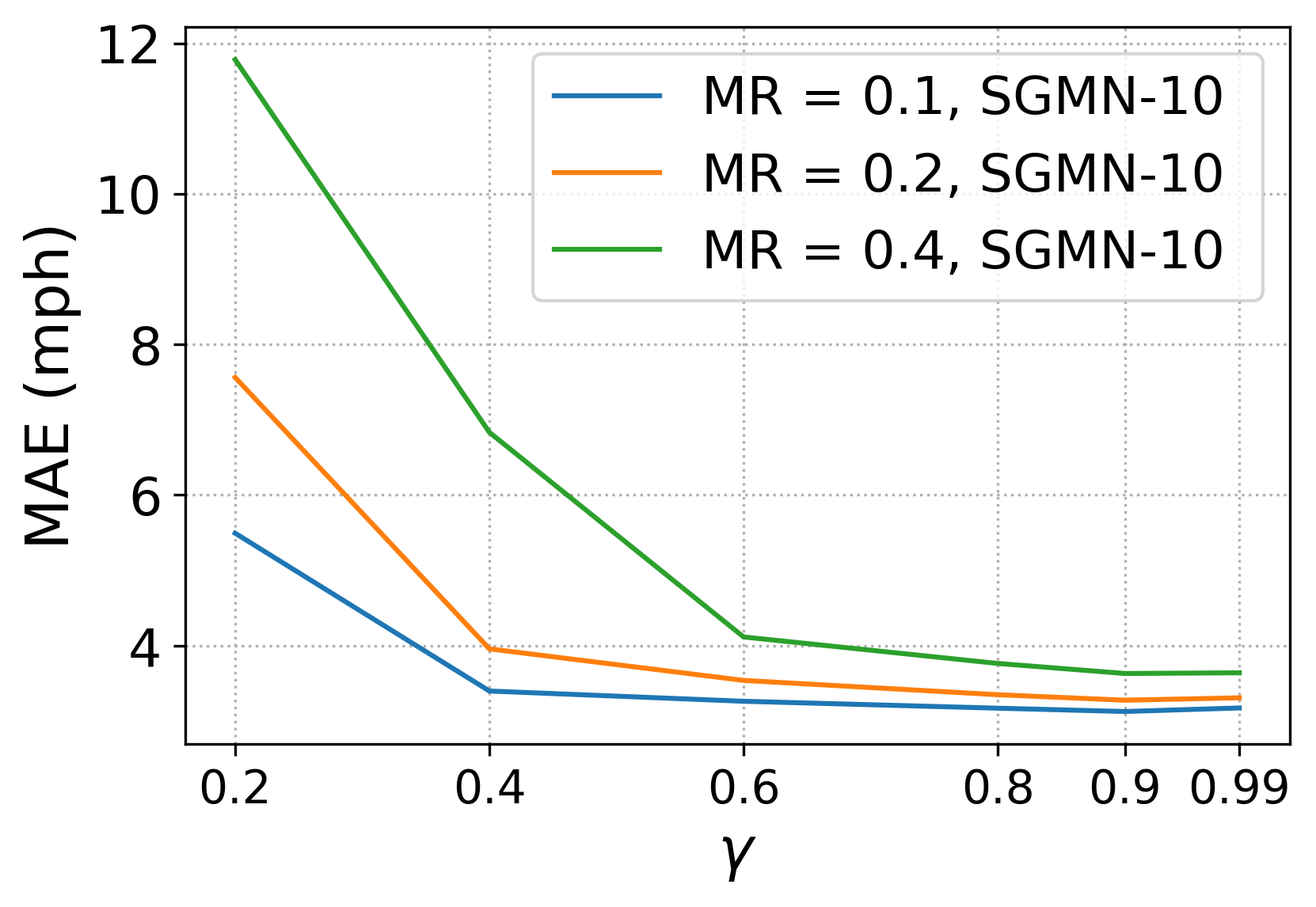}
         \caption{SGMN-10 on METR-LA}
         \label{fig:gamma_sgmn_metr}
     \end{subfigure}
     \hfill
     \begin{subfigure}[b]{0.32\textwidth}
         \centering
         \includegraphics[width=\textwidth]{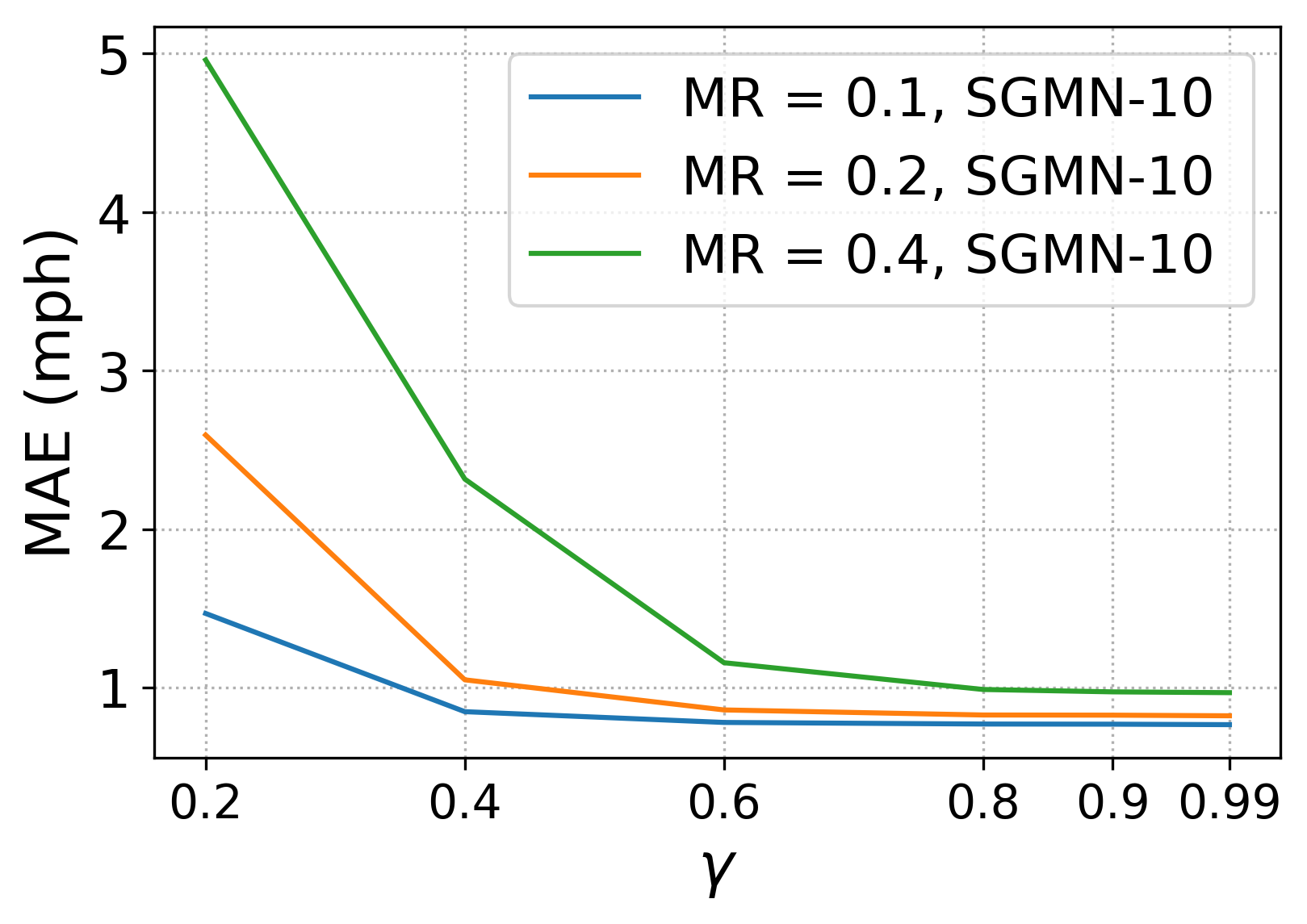}
         \caption{SGMN-10 on INRIX-SEA}
         \label{fig:gamma_sgmn_inrix}
     \end{subfigure}
     \\
     \begin{subfigure}[b]{0.32\textwidth}
         \centering
         \includegraphics[width=\textwidth]{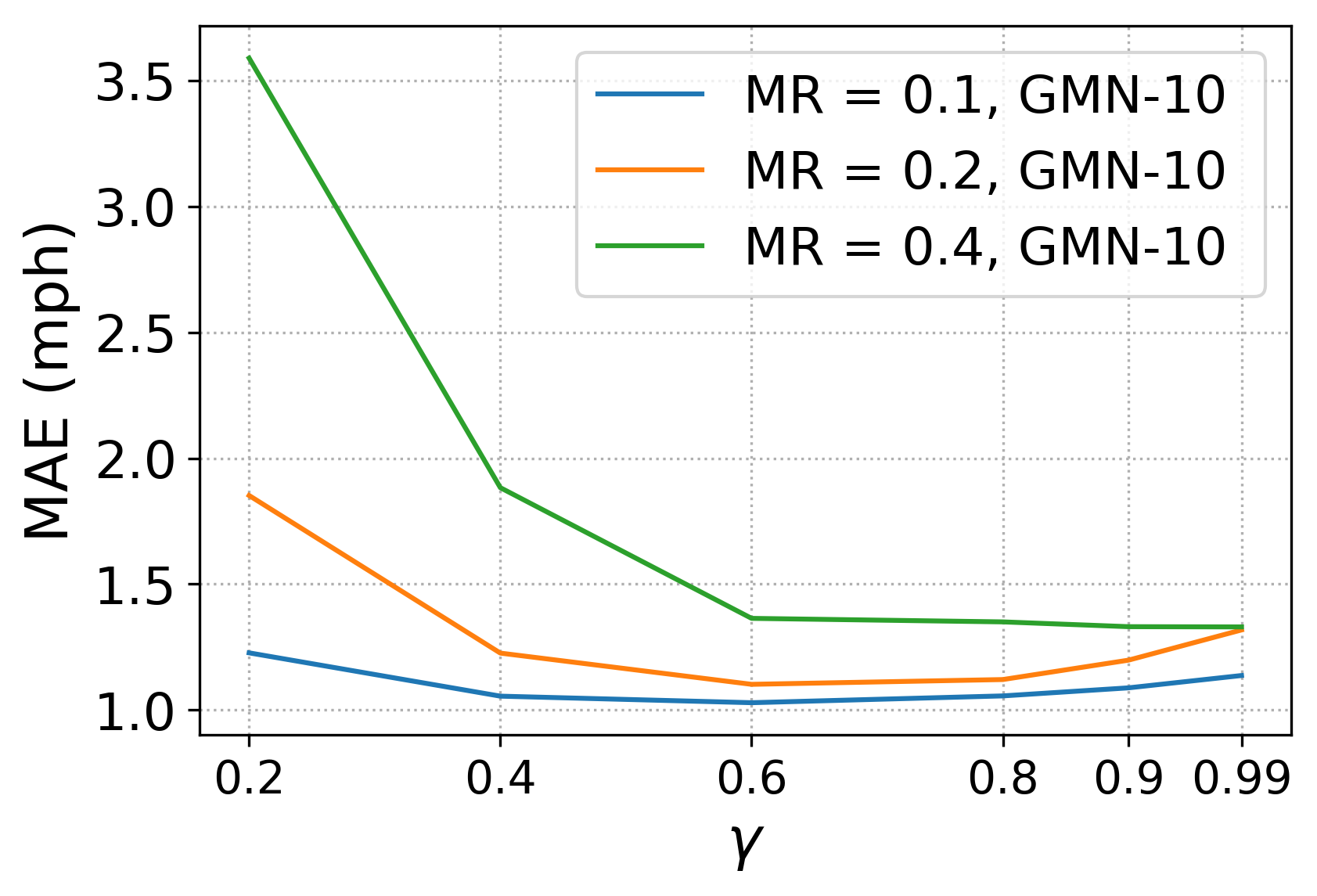}
         \caption{GMN-10 on PEMS-BAY}
         \label{fig:gamma_gmn_pems}
     \end{subfigure}
     \hfill
     \begin{subfigure}[b]{0.32\textwidth}
         \centering
         \includegraphics[width=\textwidth]{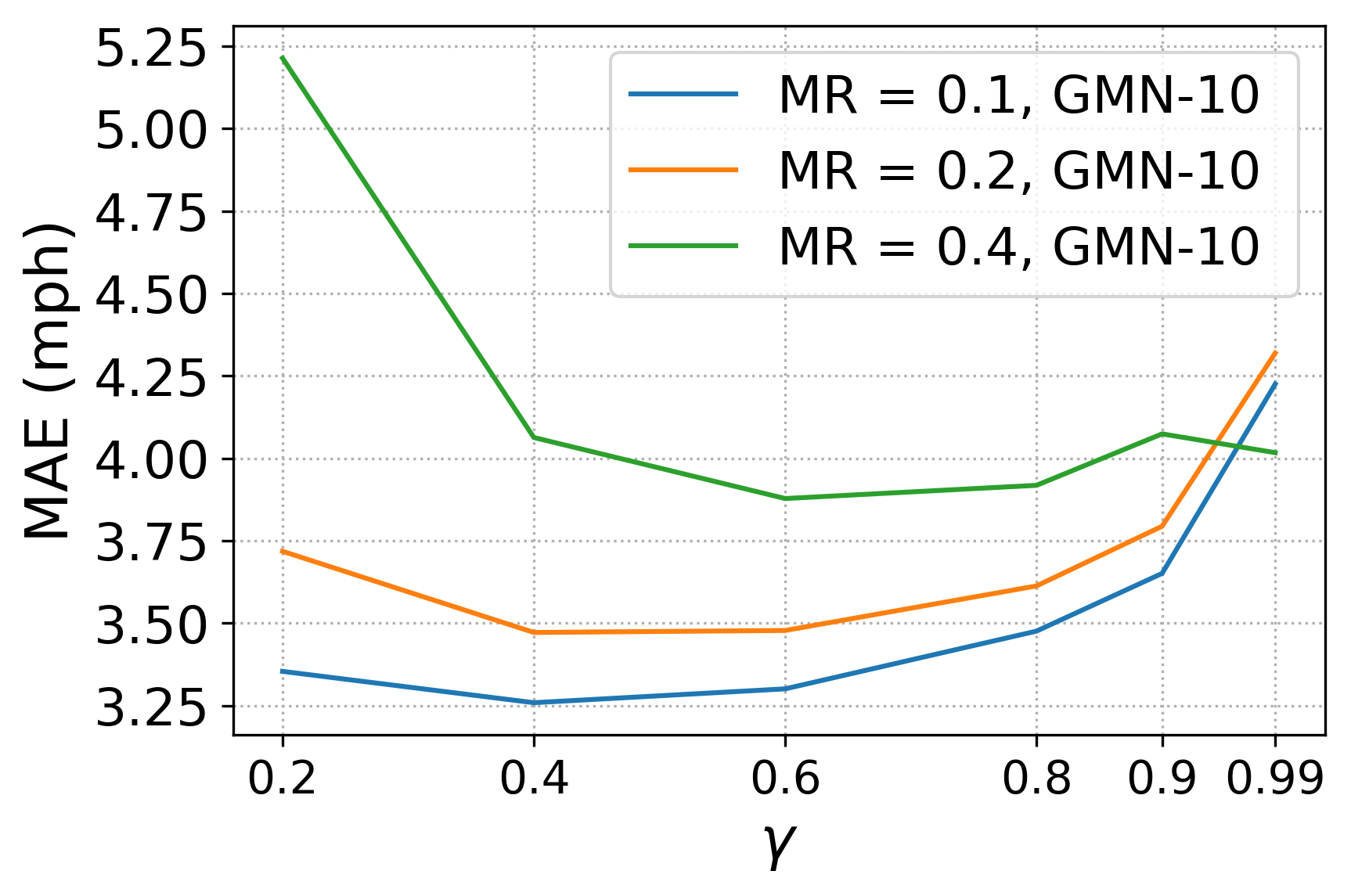}
         \caption{GMN-10 on METR-LA}
         \label{fig:gamma_gmn_metr}
     \end{subfigure}
     \hfill
     \begin{subfigure}[b]{0.32\textwidth}
         \centering
         \includegraphics[width=\textwidth]{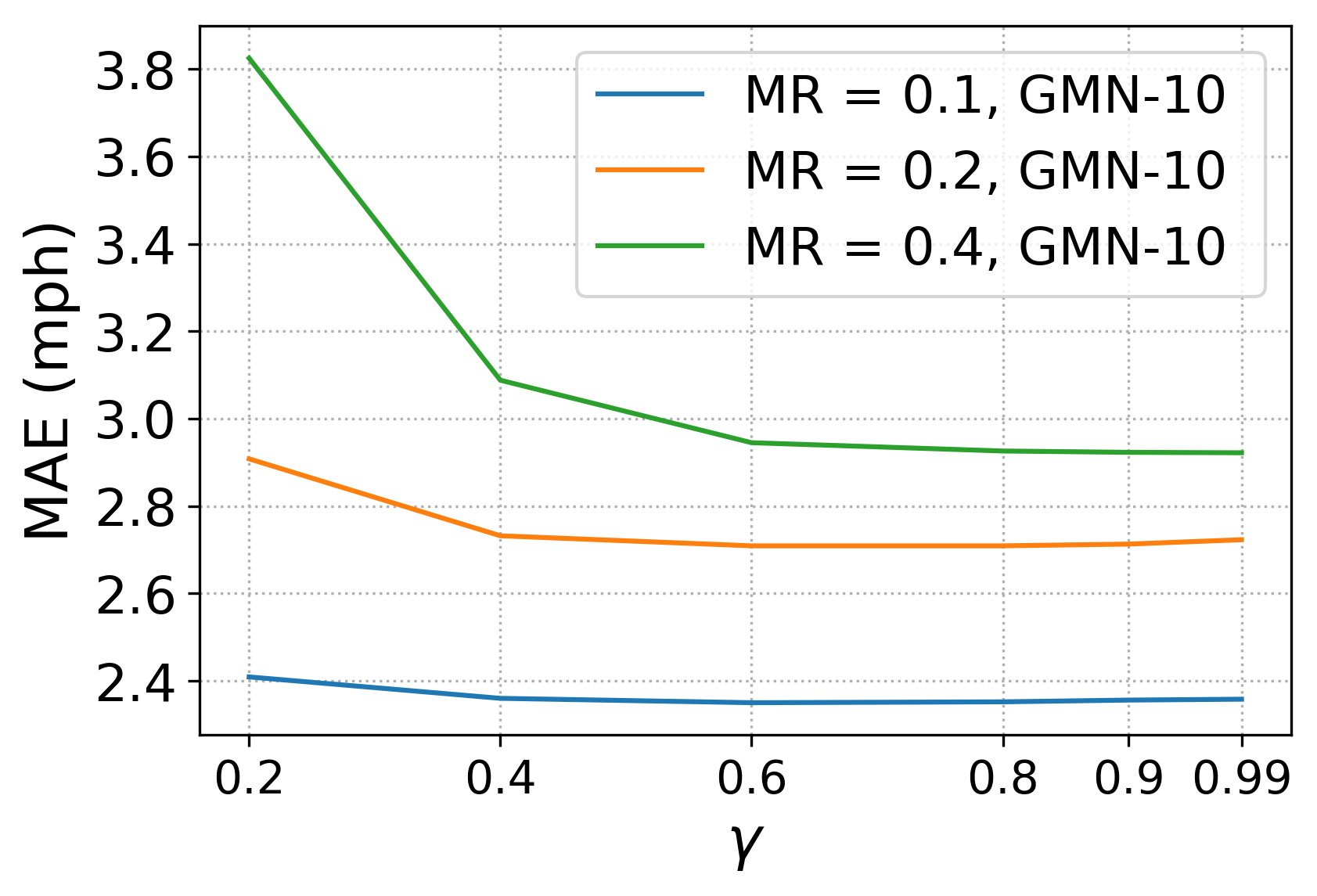}
         \caption{GMN-10 on INRIX-SEA}
         \label{fig:gamma_gmn_inrix}
     \end{subfigure}
    \caption{Prediction performance metric (MAE) w.r.t. the decay rate $\gamma$. The SGMN-10 and GMN-10 are tested on the PEMS-BAY, METR-LA, and INRIX-SEA datasets with different missing rates. }
    \label{fig:gamma}
\end{figure}

The proposed graph Markov process adopts the decay rate $\gamma \in (0,1)$ to represent the temporal impact of $\Delta t$ on the traffic state transition process. In previous analysis sections, the $\Delta t$ is 5-minutes, and the decay rates of GMN models are set as 0.9. In this section, we analyze the influence of the decay rate on the proposed models' prediction performance. The prediction performance of SGMN-10 and GMN-10 w.r.t. various decay rates are shown in Figure \ref{fig:gamma}. The models are tested on the three datasets with different missing rates. Generally, the missing rate affects the prediction performance a lot that large missing rate results in large prediction errors.
\par
The six sub-figures in Figure \ref{fig:gamma} all indicate the prediction errors (MAE) decrease along with the increase of $\gamma$. The prediction results of the SGMN-10 models tested on the three datasets have the similar curve patterns, as shown by the line-charts in Figure \ref{fig:gamma_sgmn_pems}, \ref{fig:gamma_sgmn_metr}, and \ref{fig:gamma_sgmn_inrix}. When $\gamma$ is close to zero, the prediction errors are relatively large. When $\gamma$ is increasing, the prediction errors seem to be monotonically decreasing. When $\gamma$ is close to one, the curves are almost flat and prediction errors nearly keep the same. However, as shown in Figure \ref{fig:gamma_sgmn_metr}, the MAE tested on the METR-LA dataset increases a little bit when $\gamma$ increases from 0.9 to 0.99.
\par
The MAEs of the GMN-10 models shown in Figure \ref{fig:gamma_gmn_pems}, \ref{fig:gamma_gmn_metr}, and \ref{fig:gamma_gmn_inrix} have slightly different patterns than those of SGMN-10 models. The MAE curves are not monotonically decreasing. When $\gamma$ is close to one, the prediction errors start to increase. This phenomenon is particularly obvious in the results tested on the METR-LA dataset. 
\par
In addition, it should be noted that the y-axes of those sub-figures have various ranges. When the decay rate is relatively small (close to one), the prediction capability of GMN models is better than that of SGMN models. One possible reason is that GMN models contains more weight parameters than SGMN models.

%and Figure \ref{fig:gamma_rmse} demonstrate the MAE and RMSE with respect to the $\gamma$, respectively. In the GMN, since a traffic state $x_{t-i}$'s impact on $x_{t}$ is multiplied with $\gamma^{i+1}$ and $\gamma \in (0,1)$, the smaller the $\gamma$ is, the less $x_{t-n}$ contributes to $x_t$. When the missing rate is low, the prediction results with different decay rates are similar. However, if the missing rate is relatively high (missing rate = 0.4), the $\gamma$ needs to be large enough to ensure historical traffic states away from $x_t$ can still contribute to $x_t$. In this way, the missing values in the input state sequence can be accurately inferred to help generate prediction results. When the missing rate is low and the $\gamma$ is close to 1, the trough-shaped lines in Figure \ref{fig:gamma_rmse} indicate that there is a trade-off between $\gamma$ and the missing rate for configuring/tuning parameters for GMN models.

\subsubsection{Analysis on Forecasting Residuals}

\begin{figure}[!htb]
     \centering
     \begin{subfigure}[b]{0.32\textwidth}
         \centering
         \includegraphics[width=\textwidth]{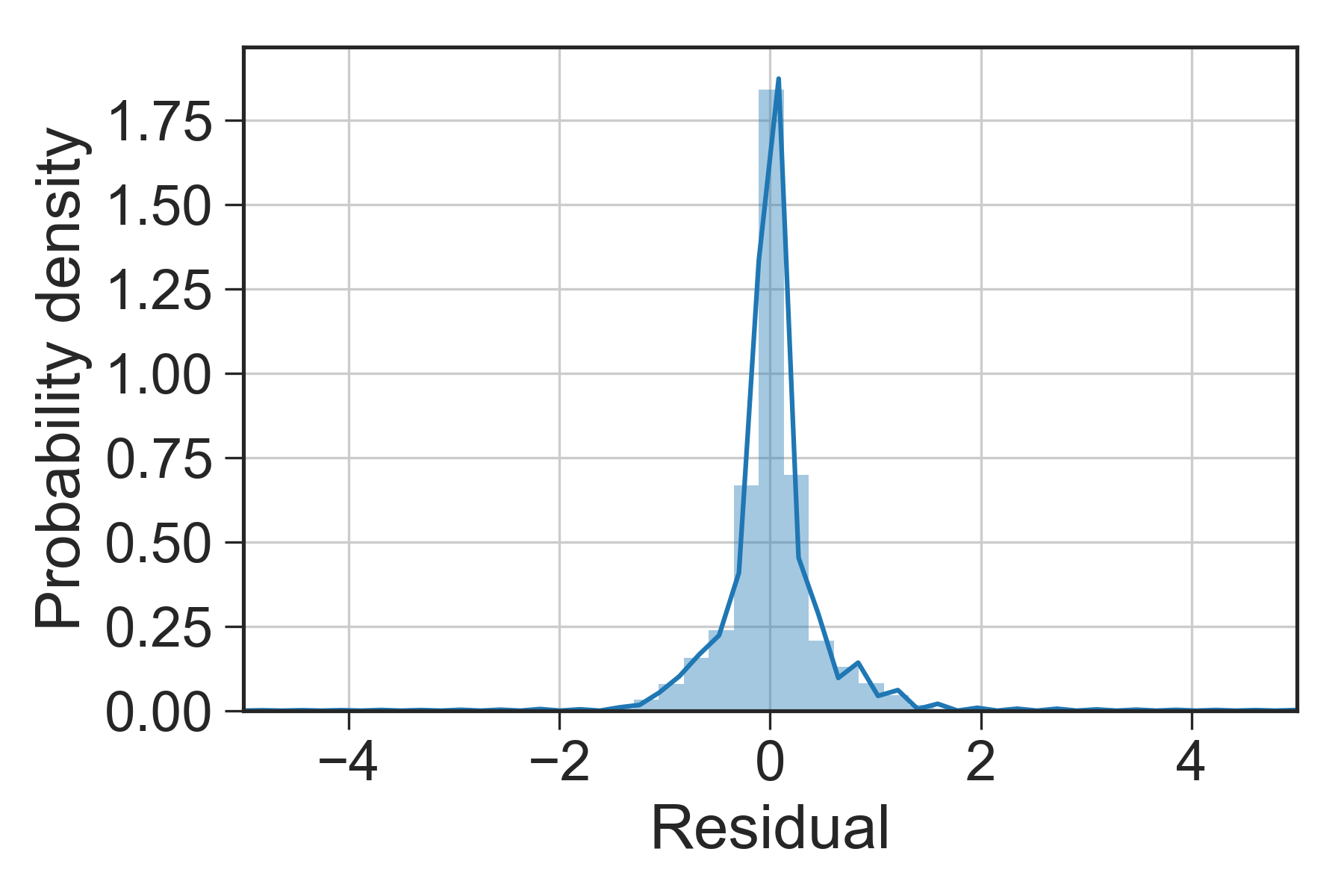}
         \caption{SGMN-10 on PEMS-BAY}
         \label{fig:residual_sgmn_pems}
     \end{subfigure}
     \hfill
     \begin{subfigure}[b]{0.32\textwidth}
         \centering
         \includegraphics[width=\textwidth]{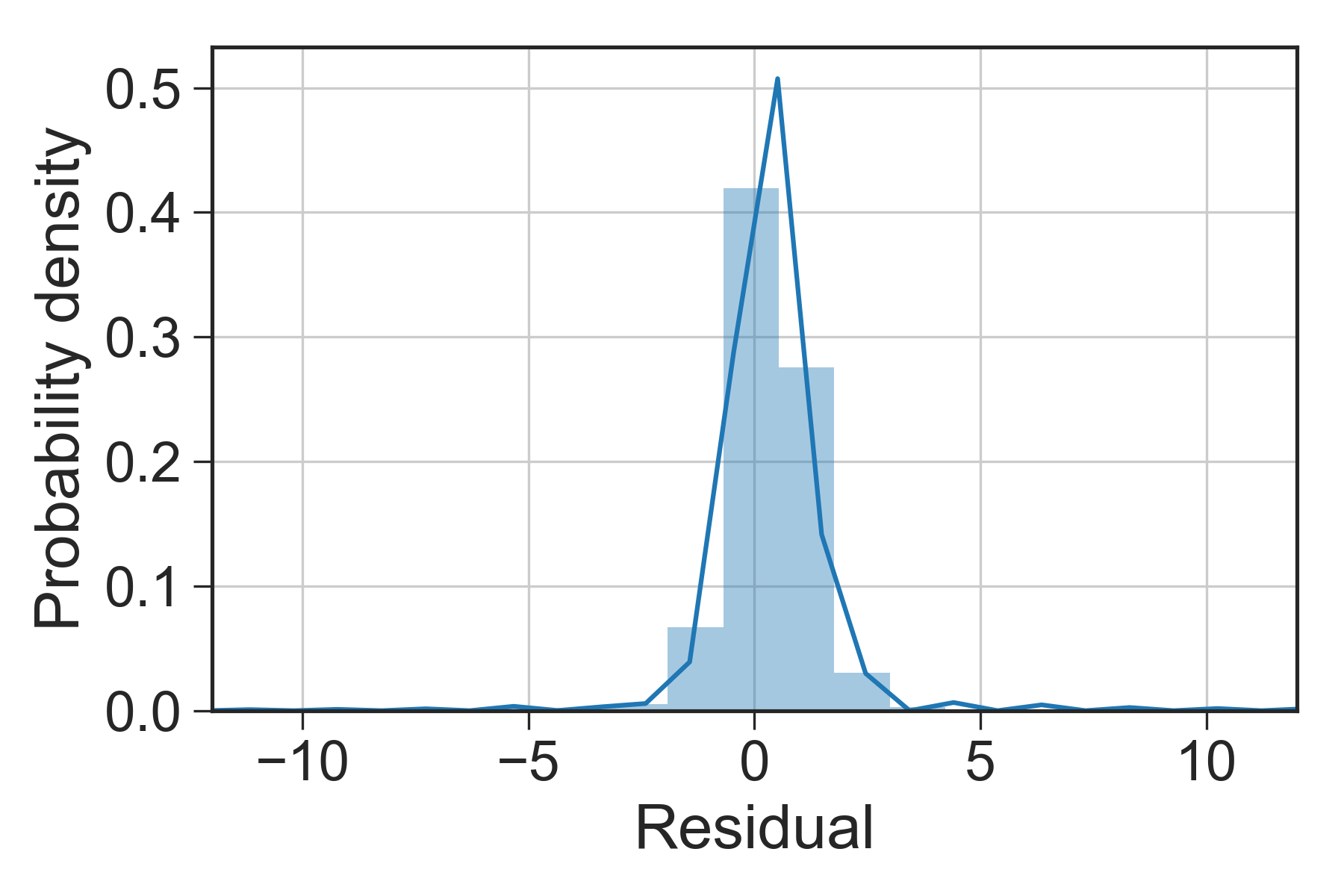}
         \caption{SGMN-10 on METR-LA}
         \label{fig:residual_sgmn_metr}
     \end{subfigure}
     \hfill
     \begin{subfigure}[b]{0.32\textwidth}
         \centering
         \includegraphics[width=\textwidth]{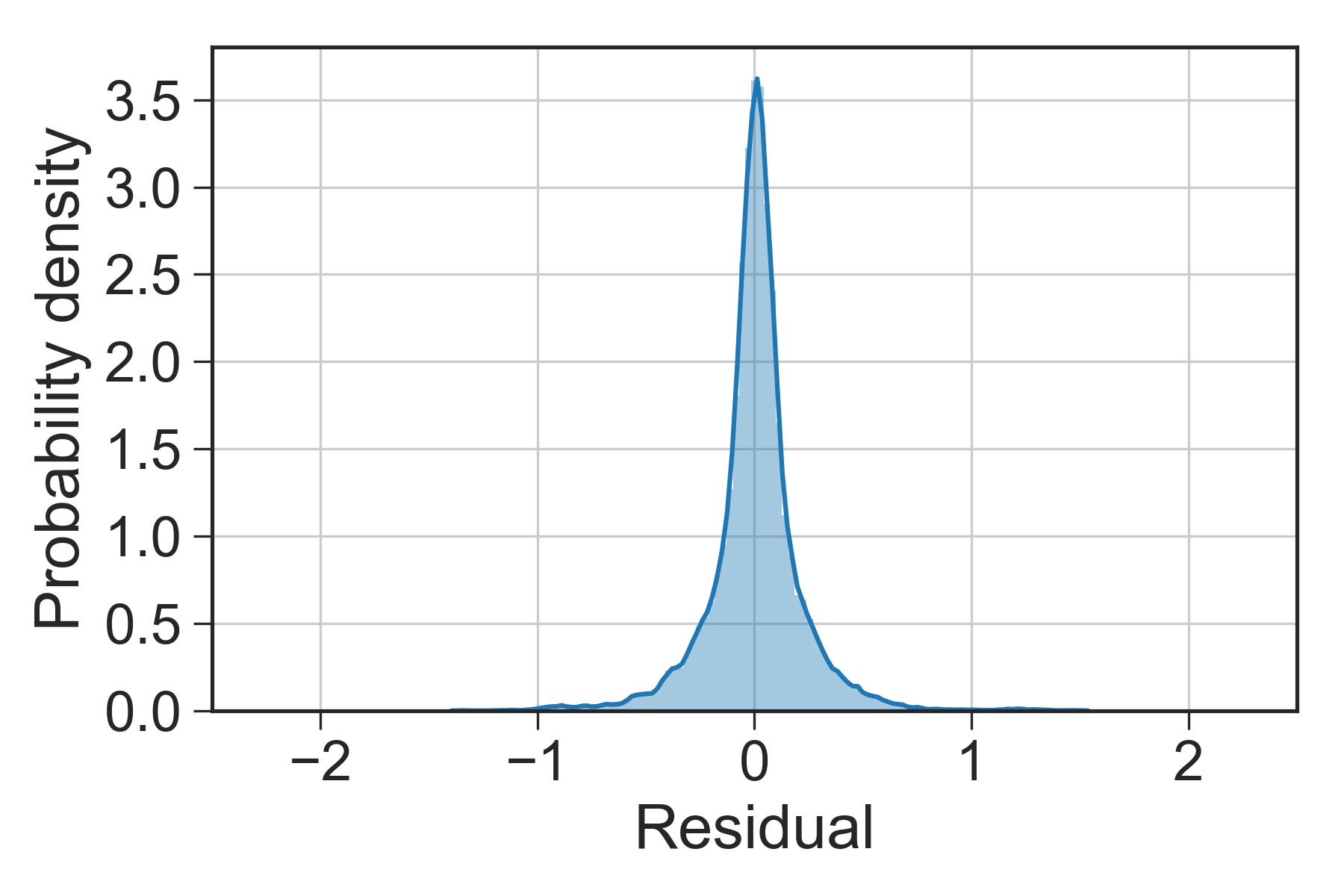}
         \caption{SGMN-10 on INRIX-SEA}
         \label{fig:residual_sgmn_inrix}
     \end{subfigure}
     \\
     \begin{subfigure}[b]{0.32\textwidth}
         \centering
         \includegraphics[width=\textwidth]{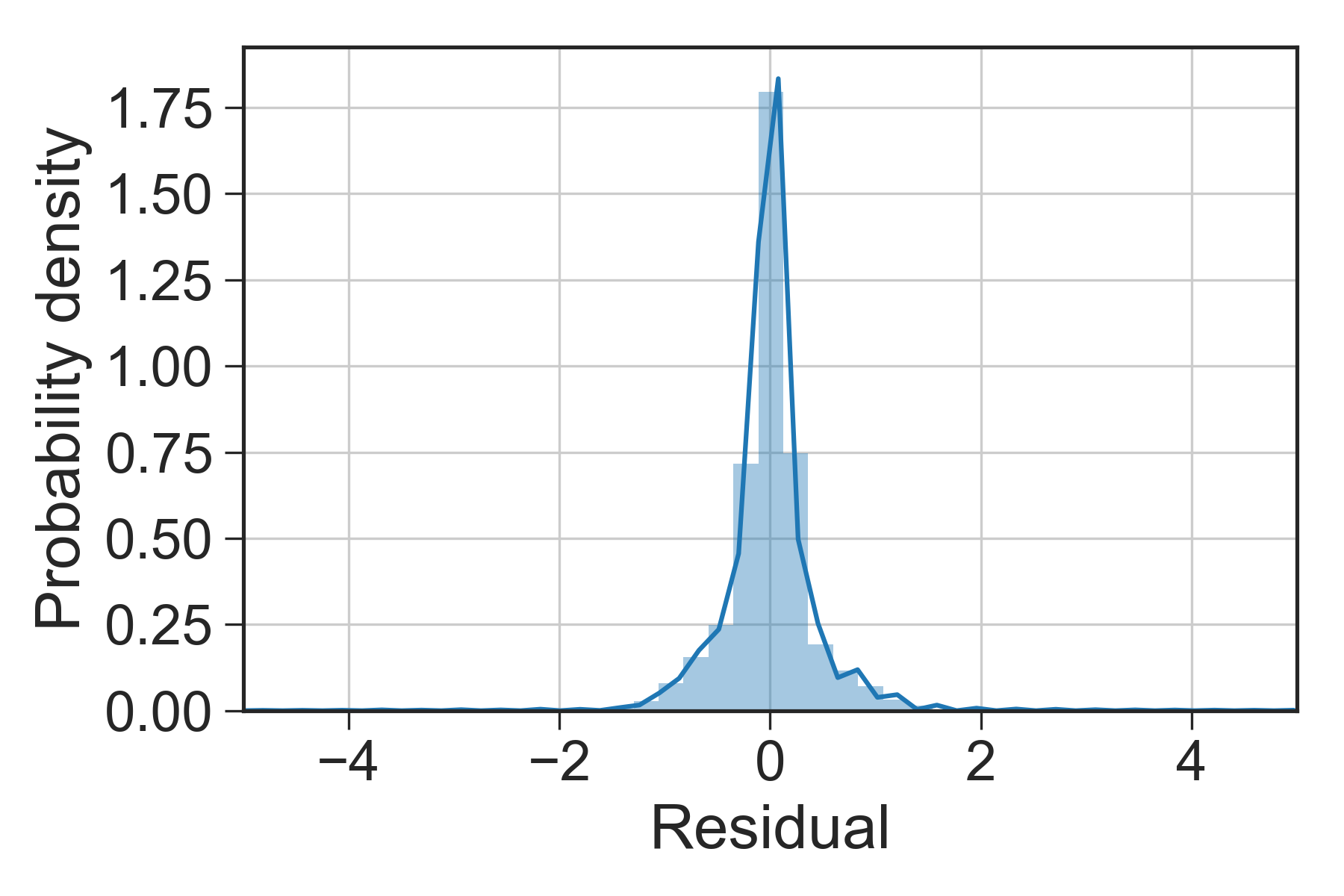}
         \caption{GMN-10 on PEMS-BAY}
         \label{fig:residual_gmn_pems}
     \end{subfigure}
     \hfill
     \begin{subfigure}[b]{0.32\textwidth}
         \centering
         \includegraphics[width=\textwidth]{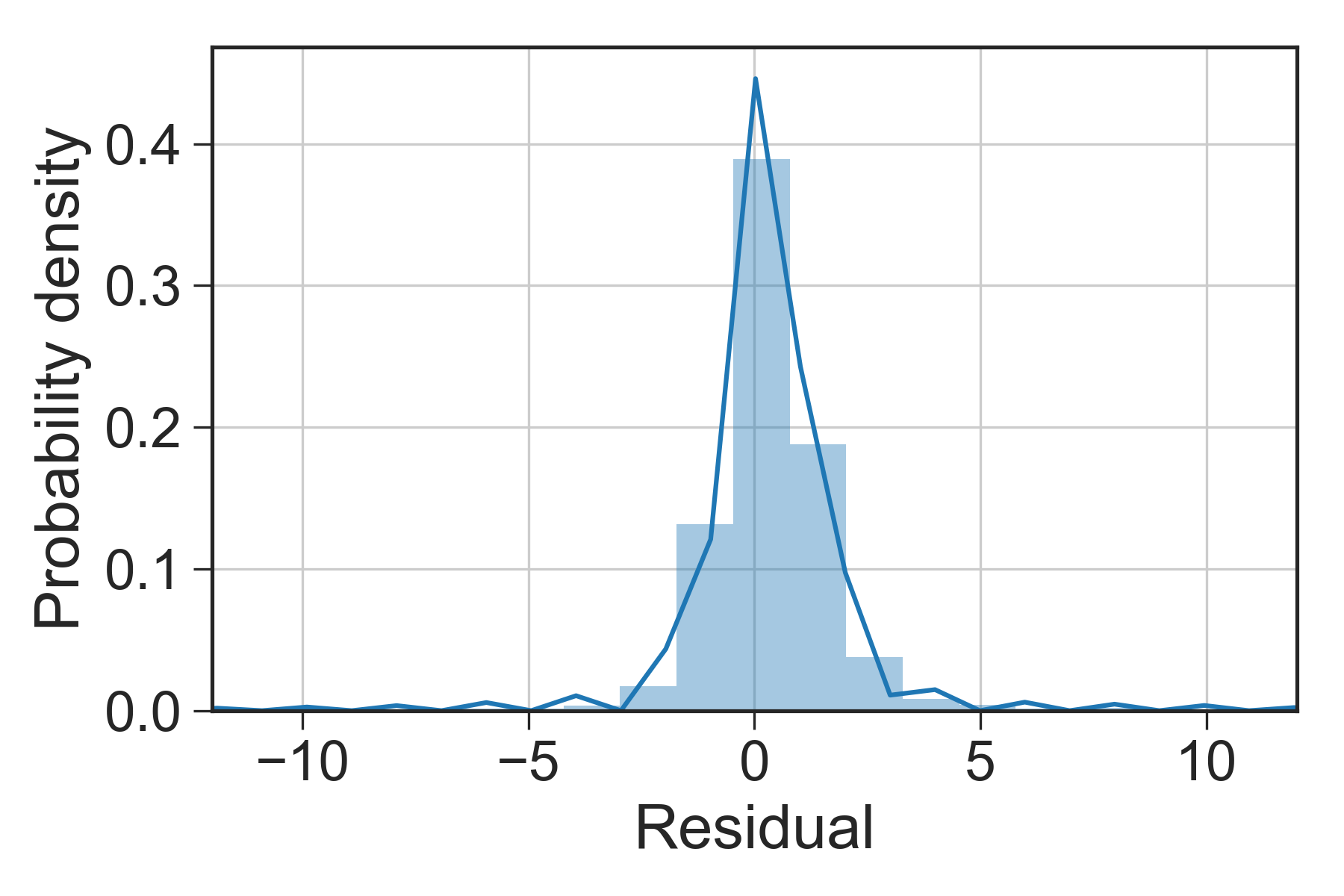}
         \caption{GMN-10 on METR-LA}
         \label{fig:residual_gmn_metr}
     \end{subfigure}
     \hfill
     \begin{subfigure}[b]{0.32\textwidth}
         \centering
         \includegraphics[width=\textwidth]{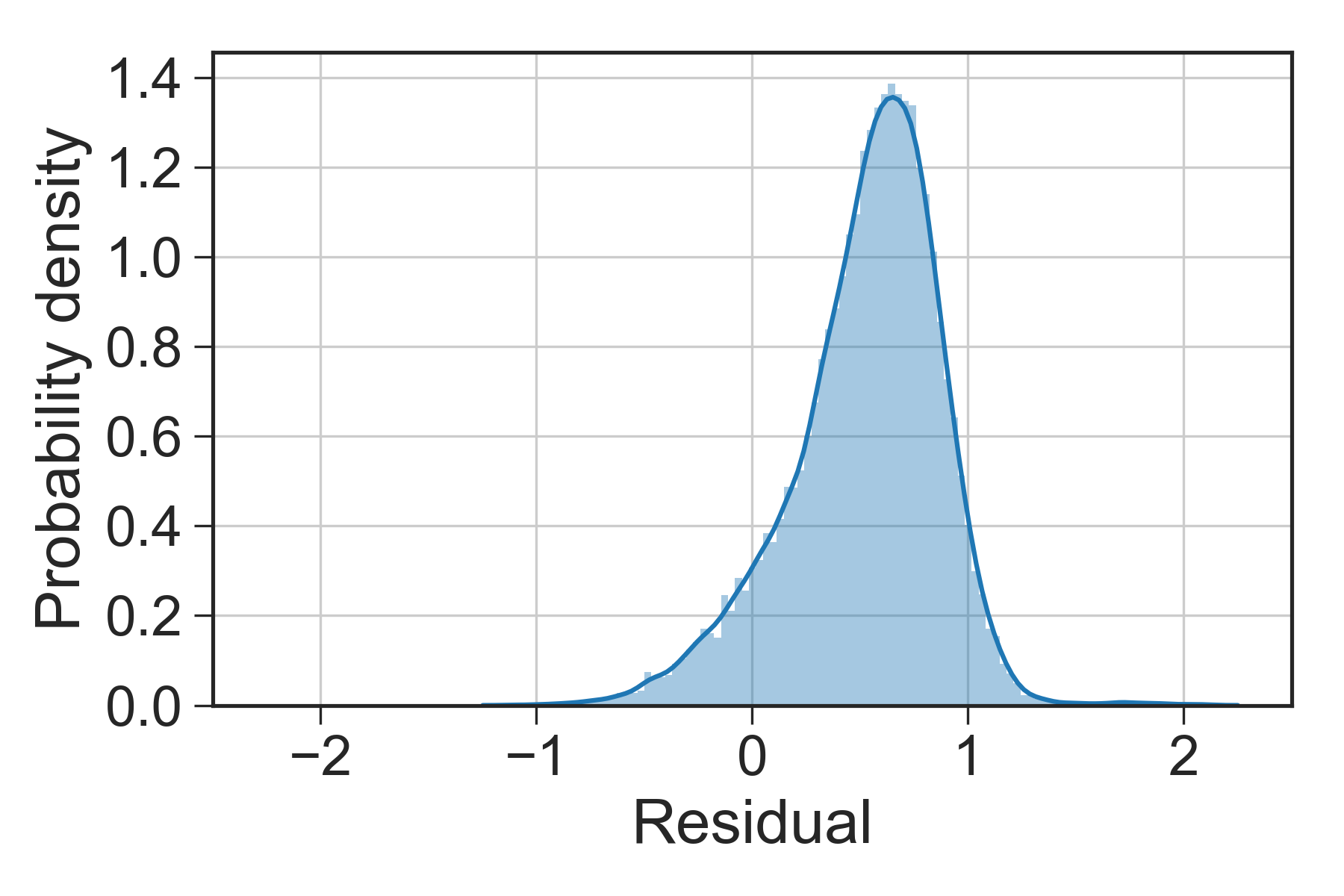}
         \caption{GMN-10 on INRIX-SEA}
         \label{fig:residual_gmn_inrix}
     \end{subfigure}
    \caption{Prediction residuals of the proposed models tested on three datasets when the missing rate is 20\%.}
    \label{fig:residual}
\end{figure}

Since residual is an critical indicator for evaluating whether a model is systematically correct, the residuals of predictions are analyzed in this section. The residual equals the ground truth value subtracts the predicted value, i.e. $x-\hat{x}$. Figure \ref{fig:residual} shows the residual distributions of SGMN-10 and GMN-10 tested on the three datasets. Most of the sub-figures display that the residual distributions follow normal distributions with zero means, except for the result of GMN-10 tested on the INRIX-SEA dataset. Although the proposed models are more complex than regression models, the residuals' normal distributions indicate that the proposed models have sufficient prediction capabilities and capture enough predictive information from the input data.

\begin{figure}[!htb]
     \centering
     \begin{subfigure}[b]{0.48\textwidth}
         \centering
         \includegraphics[width=\textwidth]{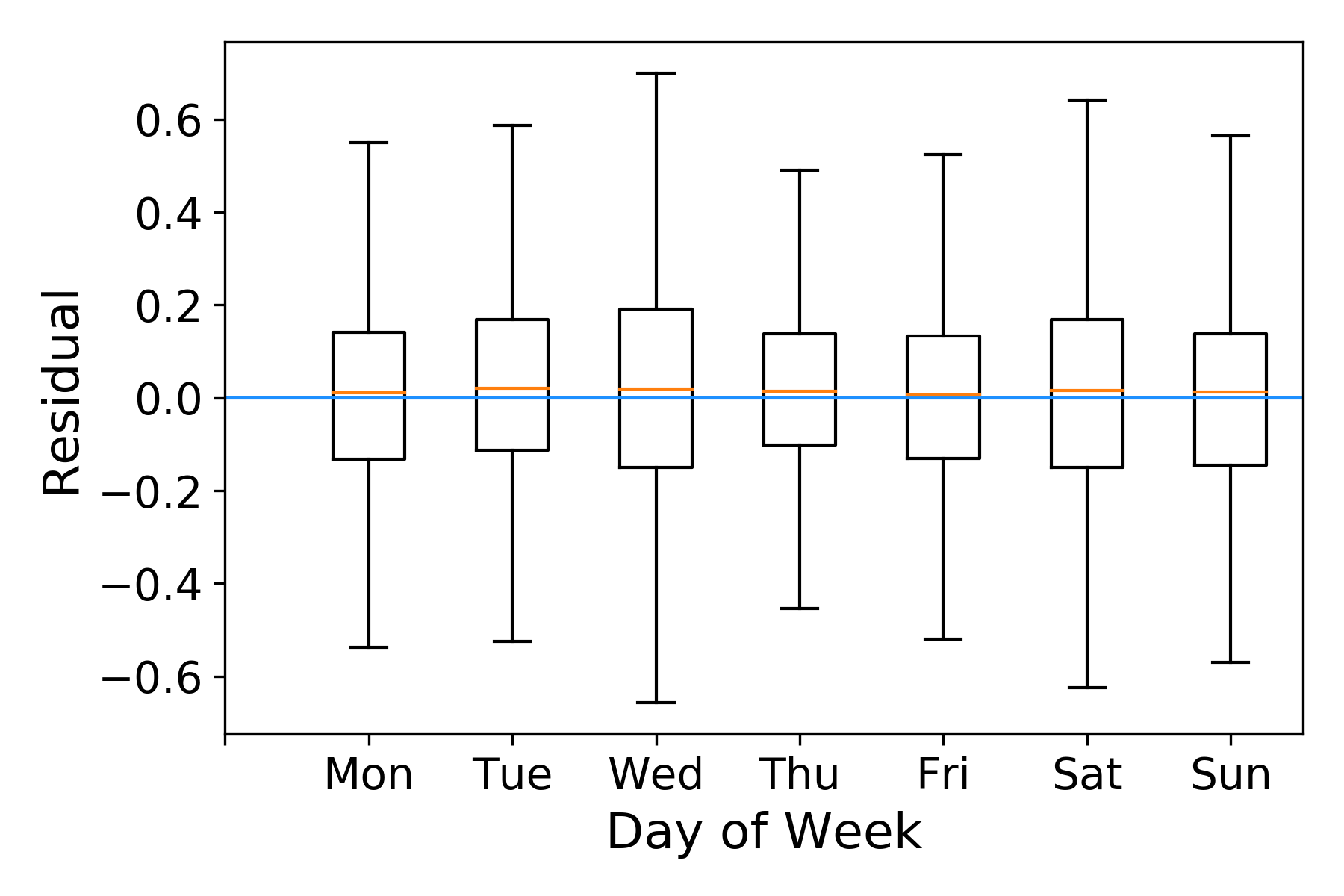}
         \caption{PEMS-BAY}
         \label{fig:residual_dow_pems}
     \end{subfigure}
     \hfill
     \begin{subfigure}[b]{0.48\textwidth}
         \centering
         \includegraphics[width=\textwidth]{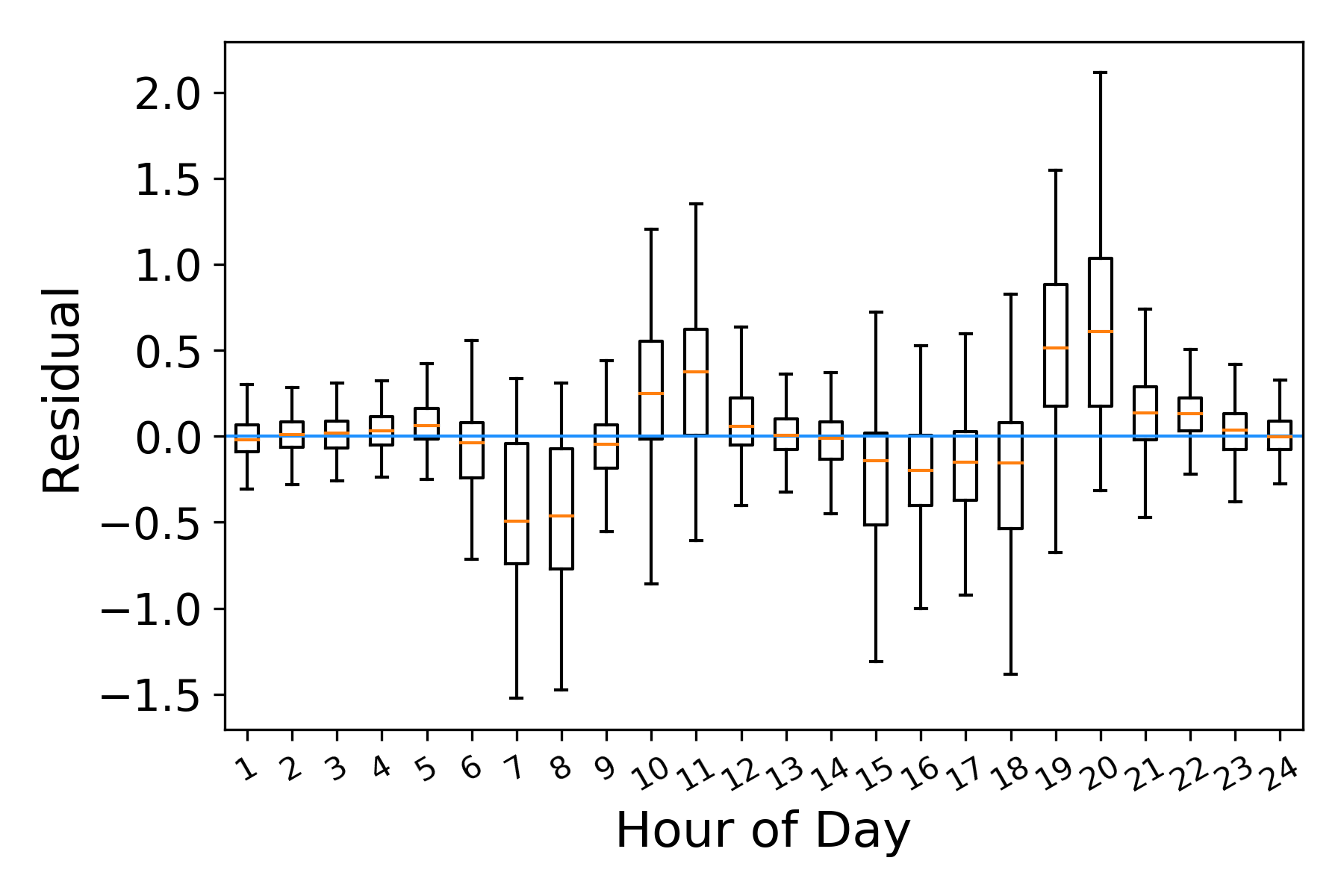}
         \caption{PEMS-BAY}
         \label{fig:residual_hod_pems}
     \end{subfigure}
     \\
     \begin{subfigure}[b]{0.48\textwidth}
         \centering
         \includegraphics[width=\textwidth]{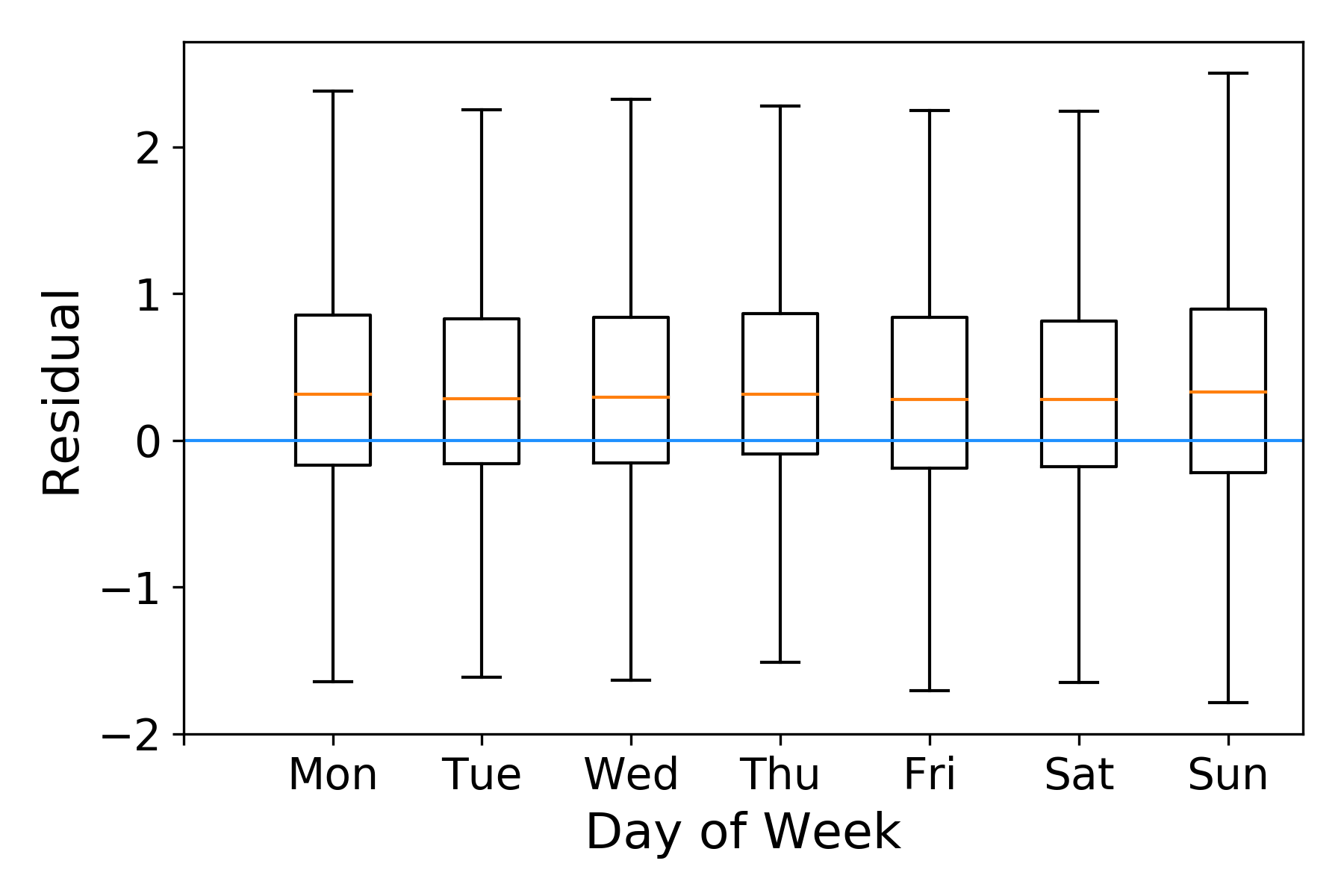}
         \caption{METR-LA}
         \label{fig:residual_dow_metr}
     \end{subfigure}
     \hfill
     \begin{subfigure}[b]{0.48\textwidth}
         \centering
         \includegraphics[width=\textwidth]{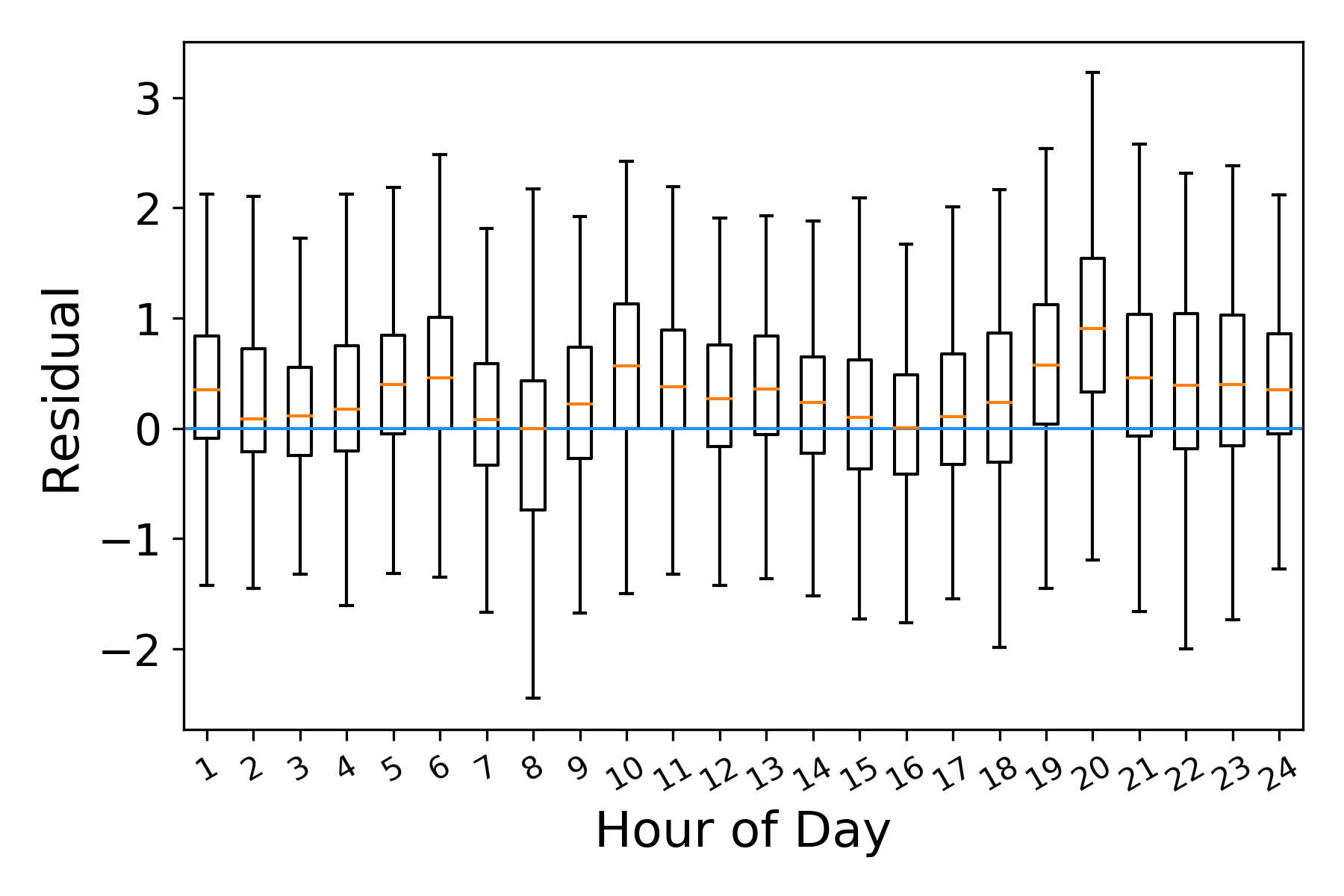}
         \caption{METR-LA}
         \label{fig:residual_hod_metr}
     \end{subfigure}
    \\
     \begin{subfigure}[b]{0.48\textwidth}
         \centering
         \includegraphics[width=\textwidth]{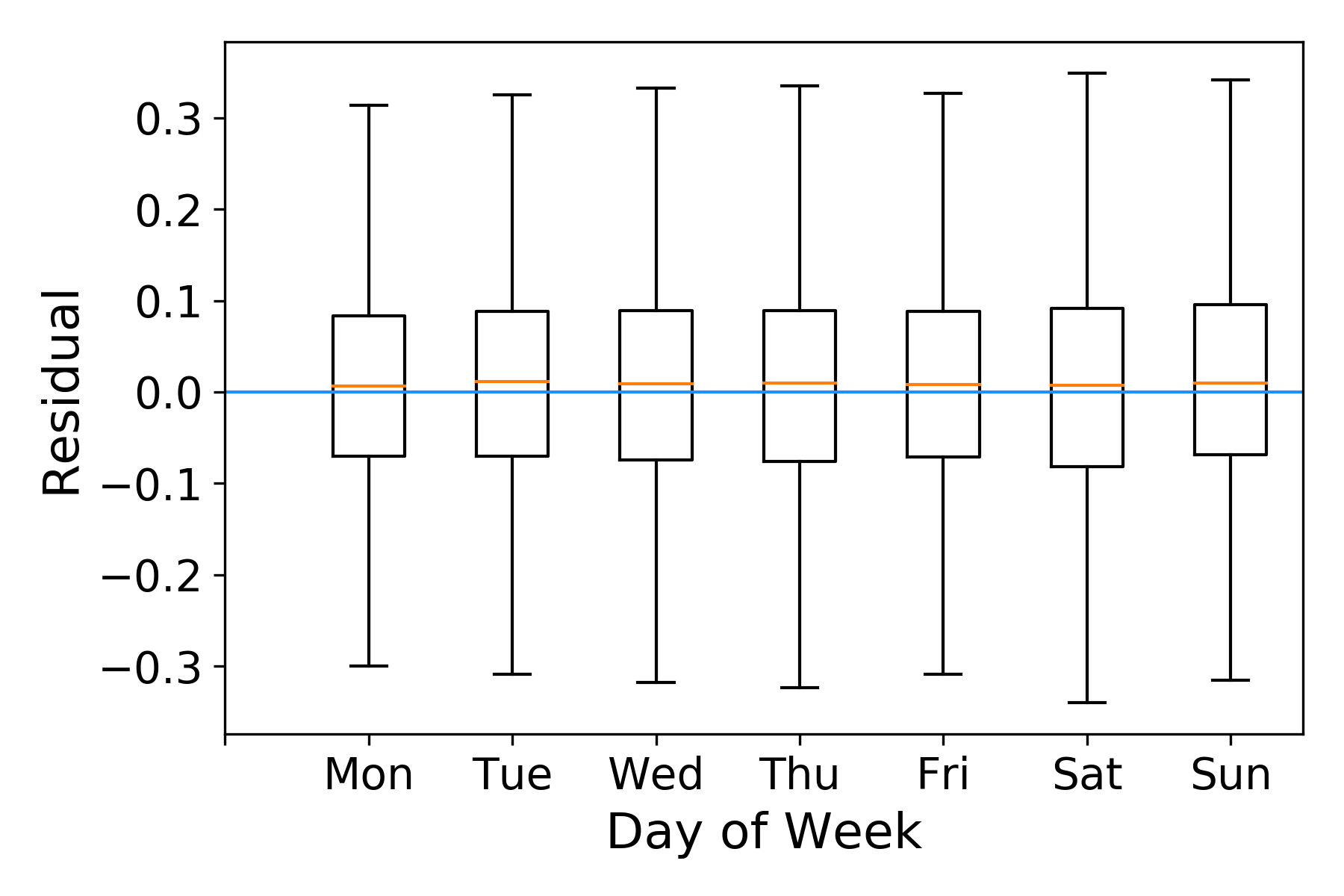}
         \caption{INRIX-SEA}
         \label{fig:residual_dow_inrix}
     \end{subfigure}
     \hfill
     \begin{subfigure}[b]{0.48\textwidth}
         \centering
         \includegraphics[width=\textwidth]{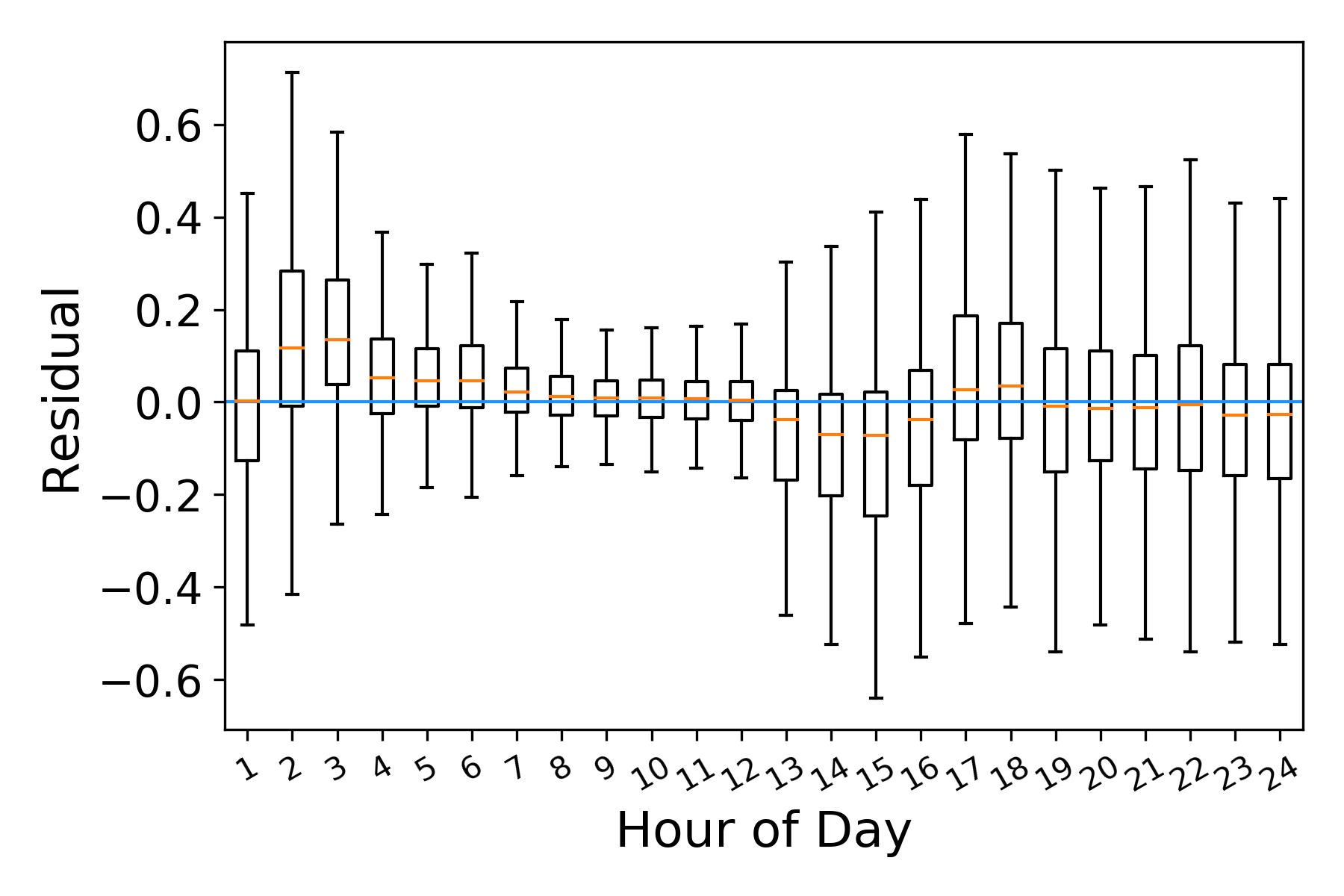}
         \caption{INRIX-SEA}
         \label{fig:residual_hod_inrix}
     \end{subfigure}
     
    \caption{Prediction residuals of SGMN-10 with respect to day of the week and hour of day, tested on three datasets with the missing rate of 20\%.}
    \label{fig:residual_time}
\end{figure}

The prediction performance will also be influenced by the temporal information, such as hour of day and day of the week. Normally, during peak hours, traffic states with more variations are harder to be predicted. Thus, in this section, the influence of hour of day and day of the week is measured. The residuals of SGMN-10 tested on the three datasets with respect to day of the week and hour of day are shown in Figure \ref{fig:residual_time}. As displayed by the box-plots in Figures \ref{fig:residual_dow_pems}, \ref{fig:residual_dow_metr}, \ref{fig:residual_dow_inrix}, the prediction residuals on each day of the week are almost the same. That means the proposed models has the capability of forecasting traffic states on each day of the week. The residuals with respect to hour of day are displayed in Figures \ref{fig:residual_hod_pems}, \ref{fig:residual_hod_metr}, \ref{fig:residual_hod_inrix}. The influence of peak hours on traffic forecasting is particularly obvious on the PEMS-BAY dataset. However, the residual distributions in each hour of the day on the METR-LA dataset do not have much differences. The residual distributions on the INRIX-SEA dataset are abnormal to some extent that the the residuals are large during the afternoon and midnight. This phenomenon may be lead by the various traffic patterns of different types of roadways in different cities.

\subsubsection{Model Weight Analysis and Visualization} \label{weight_analysis}

\begin{figure}[!htb]
     \centering
     \begin{subfigure}[b]{0.48\textwidth}
         \centering
         \includegraphics[width=\textwidth]{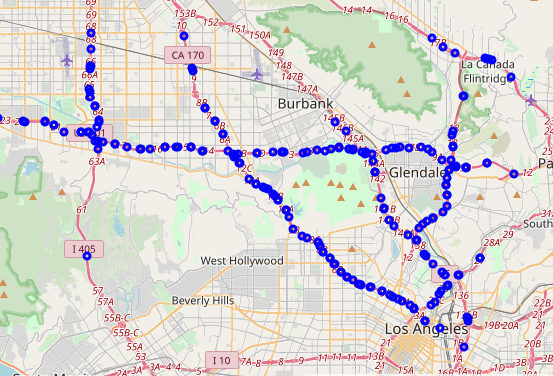}
         \caption{All sensor locations in the METR-LA dataset}
         \label{fig:sensor_location}
     \end{subfigure}
     \hfill
     \begin{subfigure}[b]{0.48\textwidth}
         \centering
         \includegraphics[width=\textwidth]{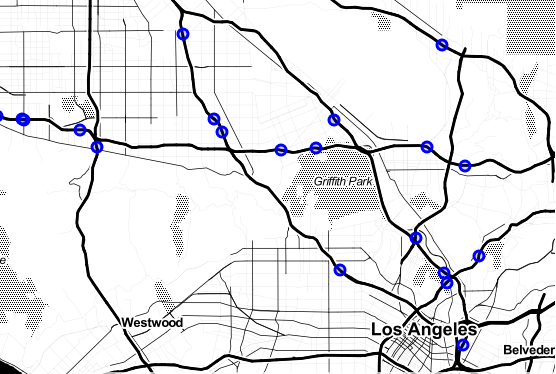}
         \caption{Top 20 influential sensor locations}
         \label{fig:top_20_nodes}
     \end{subfigure}
     \\
     \begin{subfigure}[b]{0.48\textwidth}
         \centering
         \includegraphics[width=\textwidth]{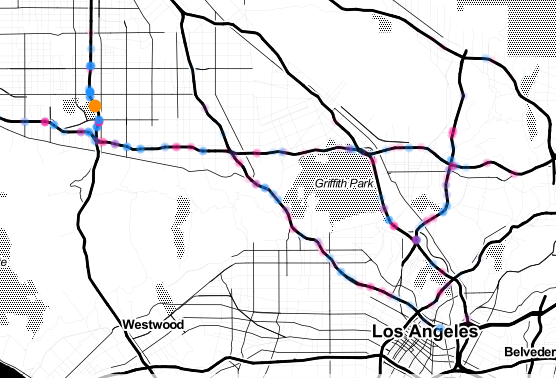}
         \caption{Weight ($U \Uplambda_{1} U^T$) visualization of SGMN-10  w.r.t. the ${89}^{th}$ sensor location, denoted by the orange dot}
         \label{fig:weight_sgmn}
     \end{subfigure}
     \hfill
     \begin{subfigure}[b]{0.48\textwidth}
         \centering
         \includegraphics[width=\textwidth]{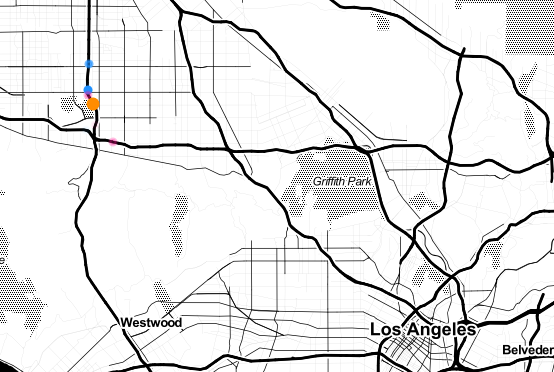}
         \caption{Weight ($\mathbf{A}^{1} \odot W_{1}$) visualization of GMN-10  w.r.t. the ${89}^{th}$ sensor location, denoted by the orange dot}
         \label{fig:weight_gmn}
     \end{subfigure}
    \caption{Visualization of sensor locations and models weights. The top 20 influential sensor locations in (b) are ones with the top 20 largest row-wise averaged squared element values of the weight matrix $H_1=U \Uplambda_{1} U^T$ in SGMN-10, which is introduced in Section \ref{weight_analysis}. The blue and pink dots in (c) and (d) represent positive and negative weight values, respectively. The darker the color is, the larger the absolute value of the weight is.}
    \label{fig:weight}
\end{figure}

In this section, the proposed model's weights are analyzed and visualized. We take the SGMN-10 and GMN-10 trained on METR-LA dataset as an example. Figure \ref{fig:sensor_location} shows the 207 sensor locations in the METR-LA dataset denoted by blue dots, and Figure \ref{fig:top_20_nodes} shows the top 20 most influential sensor locations in terms of the influence on forecasting traffic states of the future ($t+1$) step from the states of the current ($t$) step. The influence of a sensor of the $k$-th location is reflected by the sum/average of the squared element values in the $k$-th row/column of the model's weight matrix at the $t$ step, i.e. the $H_{t}$ described in Equation \ref{eq:GMP_def}. For example, the averaged squared element values of the $k$-th row of $H_{t}$ is calculated as $\frac{1}{n} \sum_{i=0}^{n-1} {(H_{t})}_{k,i}^2$. Here, $H_{t} = \mathbf{A}^{1} \odot W_{1}$ for the GMN case, and for the SGMN case, $H_{t} = U \Uplambda_{1} U^T$. As depicted in the map in Figure \ref{fig:top_20_nodes}, the selected top 20 influential sensor locations are mostly distributed near intersection areas, which has great potential to affect nearby traffic states. Figure \ref{fig:weight_sgmn} and Figure \ref{fig:weight_gmn} display the influence of the $89$-th sensor location on its neighboring sensor locations. This sensor location with the sensor ID 767351 is represented by an orange dot on the maps. The influence is reflected by the element values of the model's weight matrix at the $89$-th row/column. The positive and negative weight values of other sensor locations are demonstrated by blue and pink colors, respectively. The darker the color is, the larger the absolute value of the weight element is. The difference between these two figures is that the illustrated neighboring locations in Figure \ref{fig:weight_gmn} are confined within a small one-hop neighboring area by the weight matrix of GMN-10, i.e. the $\mathbf{A}^{1} \odot W_{1}$. However, as shown by the two figures, the surrounding sensor locations with respect to the $89$-th sensor location (the orange dot) are obviously darker, which means the traffic state of a location is influenced more by the states of its neighbors. Thus, by quantitatively analyzing or visualizing the weight matrices of the proposed models, the influence of nodes/locations in a traffic network on their neighbors can be measured.

% As described in Equation \ref{eq:GMP_def}, the graph Markov process contains $n$ transition weight matrices and the product of the these matrices $(\prod_{j=0}^{i} H_{t-j}) = (\prod_{j=0}^{i} \mathbf{A}^j \odot P_{t-j})$ measures the contribution of $x_{t-i}$ for generating the $\tilde{x}_t$.

\begin{figure*}[h!]
     \centering
     \begin{subfigure}[b]{\textwidth}
         \centering
         \includegraphics[width=\textwidth]{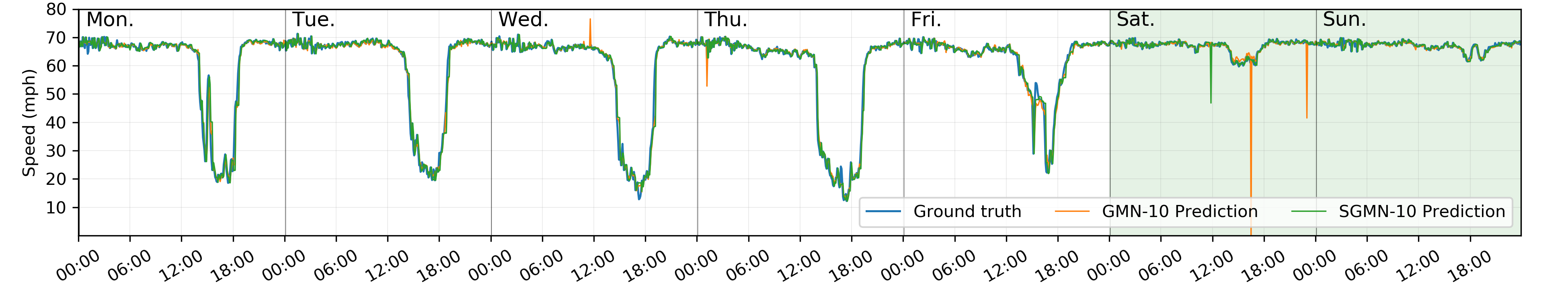}
         \caption{Results tested on the PEMS-BAY dataset from Jan. 30, 2017 to Feb. 5, 2017. Sensor ID = 400097}
         \label{fig:sensor1}
     \end{subfigure}
     \hfill\\
     \begin{subfigure}[b]{\textwidth}
         \centering
         \includegraphics[width=\textwidth]{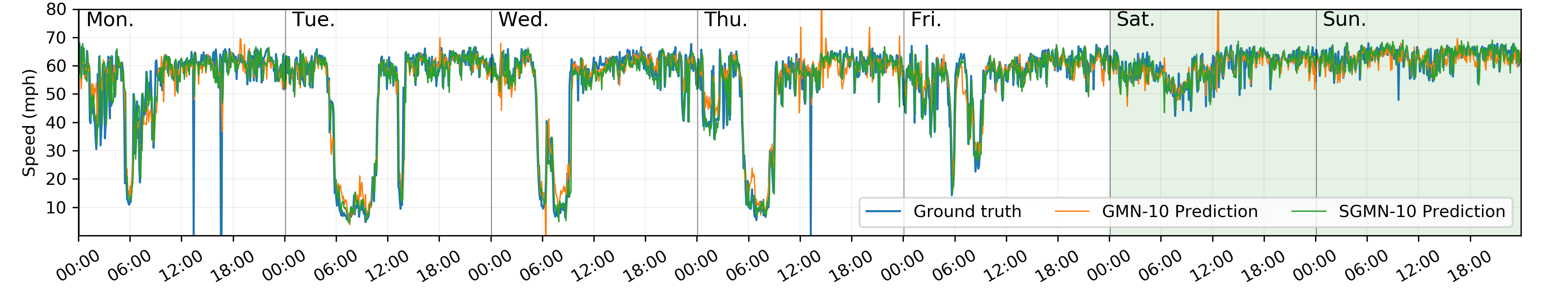}
         \caption{Results tested on the METR-LA dataset from Mar. 12, 2012 to Mar. 18, 2012. Sensor ID = 765273}
         \label{fig:sensor2}
     \end{subfigure}
     \hfill\\
     \begin{subfigure}[b]{\textwidth}
         \centering
         \includegraphics[width=\textwidth]{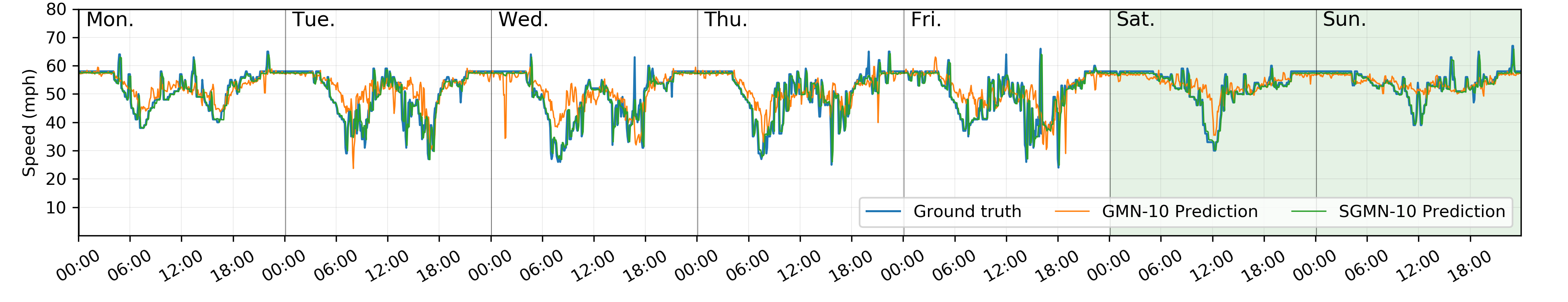}
         \caption{Results tested on the INRIX-SEA dataset from Jan. 1, 2012 to Jan. 7, 2012. link ID = 114+04549}
         \label{fig:sensor3}
     \end{subfigure}

    \caption{Comparison of the ground truth and the speed predicted by GMN-10 and SGMN-10 tested on three datasets with the missing rate of 20\% under the random missing scenario. The white and green regions in these figures demonstrate weekdays and weekends, respectively.}
    \label{fig:resultComparison}
\end{figure*}

\subsubsection{Traffic Forecasting Result Visualization}

The locations covered by those datasets actually have various traffic patterns. In this section, to demonstrate the proposed model's prediction performance, we select several sensor locations/links from the three datasets and visualize the ground truth and predicted speed values. The three sub-figures in Figure \ref{fig:resultComparison} display the ground truth and speed values predicted by the GMN-10 and SGMN-10. The missing rates of the tested datasets are all set as 20\%. Both GMN-10 and SGMN-10 work well on the PEMS-BAY dataset. Since the METR-LA dataset originally has missing values, there are some blue spikes reaching the bottom of Figure \ref{fig:sensor2} demonstrating the original missing values. The prediction performance of SGMN-10 on the INRIX-SEA dataset is better than that of GMN-10. Overall, the proposed models have the capability of forecasting traffic states with missing values.

% \subsubsection{Analysis on missing patterns}

% \subsubsection{Compare with imputation + forecasting}

% \subsubsection{simple test: add ramp to prove unclosed traffic network}

% \subsection{Experimental Results}

\section{Conclusion}

 In this study, we propose the GMN, which is a new neural network architecture for spatial-temporal data forecasting. We introduce two properties of the traffic state transition process and define a graph Markov process. Unlike other existing recurrent neural network (RNN)-based models dealing with traffic data as multivariate time series, the GMN handling the traffic state transition process as a graph Markov process. The proposed GMN can incorporate the spatial relationship between neighboring links and the links' temporal dependencies between different time steps. By incorporating the spectral graph convolution operation, we also propose a spectral graph Markov network (SGMN). The experimental results tested on a real-world dataset shows show that the proposed GMN and SGMN achieves superior prediction performance. Further, the proposed models' parameters, weights, and prediction residuals are discussed and visualized.
 
 The future work will focus on enhancing the theoretical basis of the proposed graph Markov process. We will attempt to build a connection between the graph Markov process with the Markov random field to analyze the hidden factors influencing traffic states. In addition, we will conduct more experiments on multiple public accessible datasets.
 
 \section{Acknowledgments}
 This work was supported by the Connected Cities with Smart Transportation (C2SMART) Tier 1 University Transportation Center with the USDOT Award No.: 69A3551747124. Thanks to Washington State Department of Transportation (WSDOT) for providing the research datasets. Thanks to Xinyu Chen for sharing the academic-drawing code on GitHub. Also, the authors would like to thank Ruimin Ke and Shuyi Yin for helpful discussions and comments.

\section{Reference}
%% References


\bibliographystyle{elsarticle-num}

\bibliography{reference}

\end{document}